\documentclass[a4paper]{article}
\usepackage{amsmath}
\usepackage{amssymb}
\usepackage{graphicx}
\usepackage{wrapfig}
\usepackage{enumitem}
\usepackage{float}
\usepackage{geometry}
\usepackage{quantikz}
\usepackage{tabularx}
\usepackage{lscape}
\usepackage{rotating}
\usepackage{array}
\usepackage{caption}
\usepackage{wrapfig}
\usepackage{hyperref}
\usepackage{multirow}
\usepackage{makecell}
\usepackage{longtable}
\usepackage{subcaption}
\usepackage{hyperref}
\usepackage{bm}
\usepackage{booktabs}
\usepackage{tikz}
\usetikzlibrary{shapes.geometric, arrows.meta, positioning}
\tikzstyle{startstop} = [rectangle, rounded corners, minimum width=3.5cm, minimum height=1.2cm,
text centered, draw=black, font=\large, text width=5cm, align=center]
\tikzstyle{process} = [rectangle, minimum width=3.5cm, minimum height=1.2cm,
text centered, draw=black, font=\large, text width=5cm, align=center]
\tikzstyle{decision} = [diamond, aspect=2, draw=black, font=\large, inner sep=1pt,
minimum height=1.6cm, text width=5cm, align=center]
\tikzstyle{arrow} = [thick, ->, >=stealth]
\newcommand{\citep}[1]{\cite{#1}}

\date{}  
\title{Quantum Inspired Encoding Strategies for Machine Learning Models: Proposing and Evaluating Instance Level, Global Discrete, and Class Conditional Representations}

\begin{document}
	
	\newgeometry{
		left=3cm, 
		right=3cm, 
	}
	
\maketitle
\vspace{-1.5cm}
\author{\begin{center}
		Minati Rath\textsuperscript{*1} and Hema Date\textsuperscript{2}\\
		minati.rath.2019@iimmumbai.ac.in, hemadate@iimmumbai.ac.in\\
		\textsuperscript{1}\textsuperscript{, }\textsuperscript{2} Department of Analytics and Decision Science, Indian Institute of Management (IIM) Mumbai, India	
	\end{center}}

\begin{abstract}
In this study, we propose, evaluate and compare three quantum inspired data encoding strategies, Instance  Level Strategy (ILS), Global Discrete Strategy (GDS) and Class Conditional Value Strategy (CCVS),  for transforming classical data into quantum data for use in pure classical machine learning models. The primary objective is to reduce high encoding time while ensuring correct encoding values and analyzing their impact on classification performance. The Instance Level Strategy treats each row of dataset independently; mimics local quantum states.  Global Discrete Value Based encoding strategy maps all unique feature values across the full dataset to quantum states uniformly. In contrast, the Class conditional Value based encoding strategy encodes unique values separately for each class, preserving class  dependent information.\\\\
We apply these encoding strategies to a classification task and assess their impact on encoding efficiency, correctness, model accuracy, and computational cost. By analyzing the trade  offs between encoding time, precision, and predictive performance, this study provides insights into optimizing quantum  inspired data transformations for classical machine learning workflows.

\end{abstract}

\section{Introduction}

Quantum computing has emerged as a promising field with the potential to revolutionize various domains, including machine learning \cite{dunjko2018machine}\cite{schuld2019quantum}\cite{khan2020machine} . While fully quantum models remain in their early stages due to hardware limitations, quantum inspired techniques have gained attention for improving classical machine learning workflows\cite{baniata2024sok}\cite{benedetti2019parameterized}. One such technique is quantum data encoding, which transforms classical data into quantum representations before feeding them into both classical machine learning models and quantum models for further processing and analysis \cite{rath2024quantum}\cite{schuld2017implementing}. However, a significant challenge in this approach is the high computational cost associated with encoding \cite{shin2023exponential}, making it crucial to explore efficient encoding strategies that balance accuracy and efficiency \cite{larose2020robust}.

In this study, we investigate three quantum  inspired data encoding strategies for transforming classical data into quantum representations while ensuring efficient integration with pure classical machine learning models. These strategies are:

Instance  Level Strategy – Encodes each data instance separately, preserving instance  specific variability.

Global Unique Value  Based Encoding – Maps all unique feature values in the dataset to quantum states uniformly, reducing redundancy in encoding.

Class  Specific Unique Value  Based Encoding – Encodes unique values separately for each class, preserving class  dependent information while potentially reducing encoding time.

The primary objective of this study is to reduce high encoding time while ensuring correct encoding values and analyzing their impact on classification performance. By applying these encoding methods to a classification task, we evaluate their impact on encoding efficiency, computational cost, and predictive accuracy. Our findings contribute to optimizing quantum  inspired data transformations for classical machine learning models, paving the way for practical applications of quantum techniques in real  world data  driven tasks.

\section{Literature Review}

Quantum data encoding plays a crucial role in quantum machine learning (QML) and Quantum models, as it defines how classical information is represented in quantum systems. Depending on the underlying quantum paradigm, encoding strategies can be broadly categorized into Discrete Quantum Computing (DQC) and Continuous  Variable (CV) Quantum Computing \cite{lloyd1999quantum}\cite{gu2009quantum} . The choice of encoding impacts computational efficiency, classification accuracy, and feasibility for near  term quantum devices \cite{andersen2015hybrid},\cite{choe2022quantum}.\\\\In the DQC paradigm, quantum information is processed using discrete quantum states, typically represented by qubits. Several encoding methods have been proposed to embed classical data into qubit  based quantum circuits\cite{schuld2020circuit}. Basis encoding is the simplest technique, where classical integers are directly mapped to computational basis states \cite{schuld2019quantum}\cite{perez2020data}. While this approach is computationally efficient, it requires an exponential number of qubits to represent large datasets. Binary encoding reduces qubit requirements by expressing classical values as binary strings, which are then mapped onto quantum registers \cite{havlicek2019supervised}. However, this method is limited by the overhead of binary  to  quantum conversion. More flexible alternatives include angle encoding and amplitude encoding. Angle encoding maps classical features to quantum rotation gates, providing a compact and scalable representation \cite{mitarai2018quantum}. Amplitude encoding, on the other hand, embeds classical vectors into quantum state amplitudes, achieving an exponential reduction in qubit requirements \cite{schuld2018supervised}. Despite its efficiency, amplitude encoding demands complex state preparation, making it challenging for current quantum hardware.\\\\ In contrast, the CV quantum paradigm utilizes continuous quantum states, often represented by quadratures of optical modes. Encoding methods in this framework take advantage of quantum optical systems to represent real valued classical data. Quadrature encoding is a fundamental technique, where classical values are encoded into position or momentum quadratures of quantum states \cite{braunstein2005quantum} \cite{lloyd2020quantum}. This method provides high precision but requires specialized quantum hardware \cite{ohliger2012efficient}. Displacement encoding, another CV approach, represents classical data through displacements in phase space, enabling efficient computation in photonic quantum processors \cite{braunstein2005quantum}. These CV encoding methods have been explored for machine learning applications, particularly in quantum  enhanced kernel methods and variational circuits \cite{zoufal2024quantum}.\\\\While these encoding techniques were initially designed for purely quantum models, recent research has investigated their application in hybrid quantum  classical frameworks and classical machine learning and quantum neural network models \cite{killoran2019continuous}\cite{banchi2021measuring}. Quantum  inspired encoding strategies, such as global feature mapping and unique value  based encoding, have been proposed to improve classical model performance \cite{benedetti2019parameterized}. However, a major challenge remains: the high computational cost of encoding, particularly for large datasets. Studies have explored various optimization techniques to reduce encoding redundancy, including instance  specific encoding and class  dependent encoding schemes \cite{grant2023encoding}. Despite these advancements, the trade  off between encoding accuracy and computational efficiency remains an open research question.\\\\This study aims to address this gap by systematically evaluating quantum  inspired encoding strategies in classical machine learning models. Specifically, we analyze   Instance  Level Strategy, global unique value  based encoding, and class  specific unique value  based encoding, assessing their computational cost and classification performance. By investigating these methods, we contribute to the development of efficient quantum  inspired data transformation techniques for classical machine learning applications.

\section{Quantum Data Encoding Strategies}
Encoding classical data into quantum representations is a crucial step in leveraging quantum computing for machine learning tasks. Unlike classical machine learning models that operate on numerical or categorical inputs, quantum models require data to be mapped onto quantum states. The choice of encoding method significantly impacts computational efficiency, expressiveness, and model performance\cite{xiong2025fundamental}. Traditional   Instance  Level Strategy preserves full instance  specific information but comes at the cost of high computational complexity. In contrast, global unique value  based encoding reduces redundancy by encoding each unique feature value only once, optimizing computational efficiency. A further refinement, class  specific unique value  based encoding, ensures that unique values are encoded separately for each class, preserving class  dependent patterns while maintaining a balance between efficiency and classification performance. The following subsections discuss these three encoding strategies, highlighting their mathematical formulations, advantages, and trade  offs in the context of quantum machine learning.

\subsection{Instance Level Strategy (ILS)}

\begin{wrapfigure}{r}{0.7\textwidth}
	\begin{center}
		\includegraphics[width=0.6\linewidth]{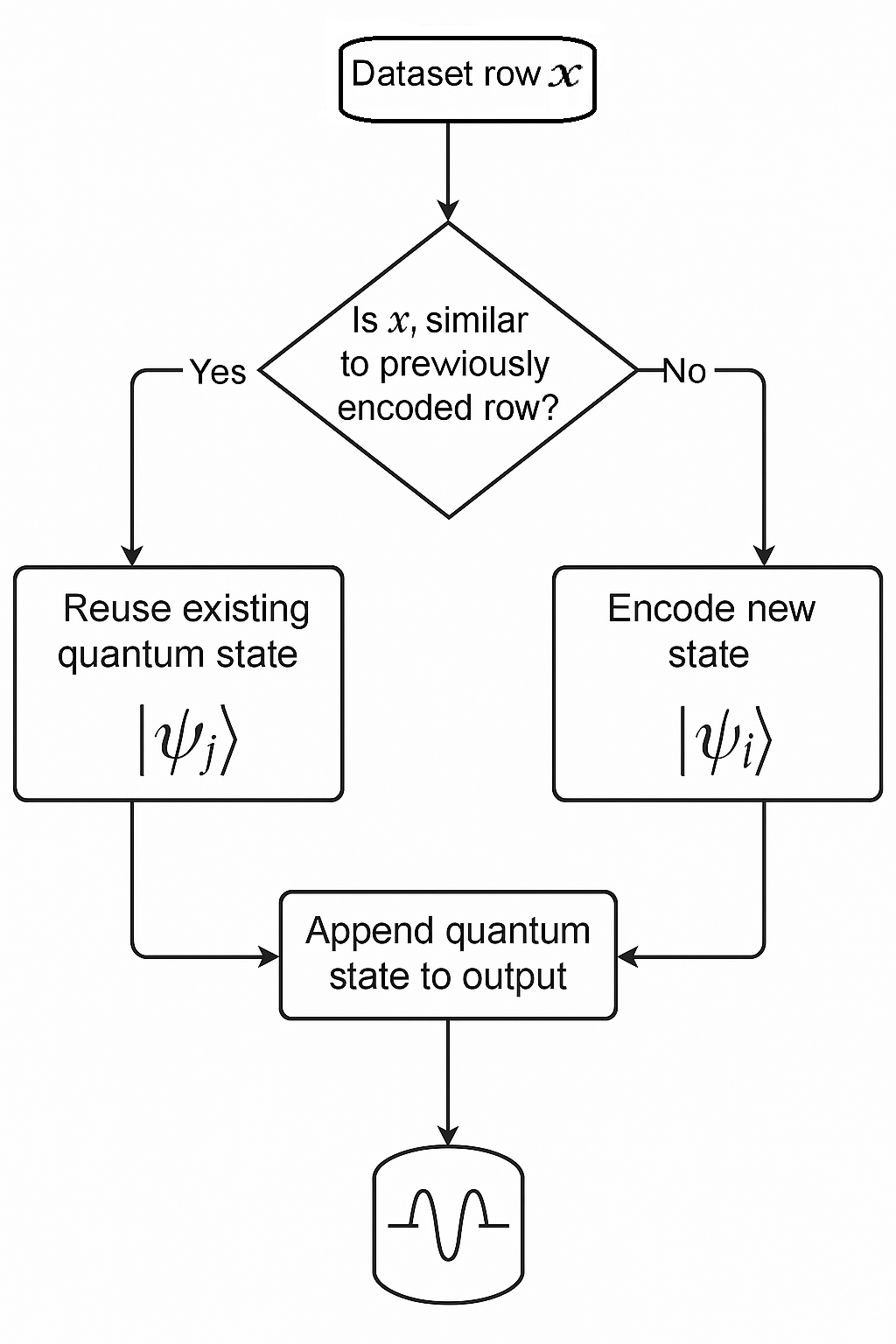}
		\caption{Flow Chart for Instance level strategy}
		\label{fig:ILSFlowChart}
	\end{center}
\end{wrapfigure}

Instance Level Strategy involves transforming each data instance individually into a quantum representation. This method preserves instance specific variability and ensures that each row is independently mapped to a quantum state. Unlike other encoding strategies that optimize redundancy, Instance  Level Strategy retains the full granularity of the dataset. \\ For a dataset $X = \{x_1, x_2, \dots, x_n\}$, where each $x_i$ is a feature vector corresponding to an instance,   Instance  Level Strategy follows:
\begin{equation}
	E(x_i) \to |\psi_i\rangle
\end{equation}

\noindent where $|\psi_i\rangle$ is the quantum representation of the entire row. This approach as depicted in Figure \ref{fig:ILSFlowChart}ensures no loss of information, as each instance retains its distinct representation. However, it comes with significant computational costs. The overall complexity of Instance Level Strategy depends on both the dataset size and the nature of the quantum encoding used. For a dataset with $n$ instances and $d$ features, the complexity is given by: $ O(n \cdot d \cdot C_{\text{embed}}) $
where $C_{\text{embed}}$ denotes the cost of embedding a single feature value into a quantum state. This term is encoding dependent and may vary based on the choice of embedding method such as angle embedding, basis embedding, amplitude embedding, or more advanced entangled or variational encodings.\\\\While the term $n \cdot d$ captures the classical traversal of the dataset, the actual quantum circuit complexity arises from $C_{\text{embed}}$. For instance, in angle embedding, each feature typically corresponds to a single rotation gate, whereas amplitude encoding may require normalization and multi-qubit gate construction, resulting in a higher $C_{\text{embed}}$.

Therefore, although ILS maintains full instance granularity, the practical cost is governed by both data size and the computational depth of the chosen embedding scheme.\\\\This method is computationally expensive, particularly for large datasets. \\\\Despite its high computational complexity, ILS incorporates an efficiency mechanism—similar instances are identified and encoded using shared quantum representations. This prevents redundant encoding and significantly reduces encoding time in practice.\\\\ILS offers maximum granularity and fidelity, making it suitable for models that benefit from fine grained patterns. The reuse of encodings for similar rows provides a balance between accuracy and computational efficiency, enabling scalable encoding without compromising uniqueness.\\
For example, \[
X = \begin{bmatrix}
	0.2 & 0.4 & 0.6 \\
	0.2 & 0.4 & 0.6 \\
	0.9 & 0.1 & 0.3 \\
\end{bmatrix}
\]
Rows 1 and 2 are identical and are both represented by $|\psi_1\rangle$. Row 3 is unique and gets a separate state $|\psi_3\rangle$.

\subsection{Global Discrete Strategy (GDS)}
This encoding strategy optimizes computational efficiency by identifying all unique feature values in the dataset, encoding them quantumly, and then mapping them back to construct the quantum dataset. This significantly reduces redundancy as the same values appearing multiple times are only encoded once. \\ Let $X = \{x_{i,j}\}$ be a dataset with $i$ instances and $j$ features. Define $U = \{u_1, u_2, \dots, u_m\}$ as the set of unique values appearing across all features. The encoding process follows these steps:

\begin{enumerate}
	\item Identify Unique Values: Extract all unique values from the dataset.
	\item Quantum Encoding: Apply the encoding function $E$ to each unique value $u_k$, mapping it to a quantum state:
	\begin{equation}
		E(u_k) \to |\psi_k\rangle, \quad \forall k \in \{1,2,\dots,m\}
	\end{equation}
	\item Reconstruct the Quantum Dataset: Replace each occurrence of $u_k$ in the dataset with its corresponding quantum state $|\psi_k\rangle$.
\end{enumerate}

\begin{wrapfigure}{r}{0.6\textwidth}
	\begin{center}
		\includegraphics[width=0.6\linewidth]{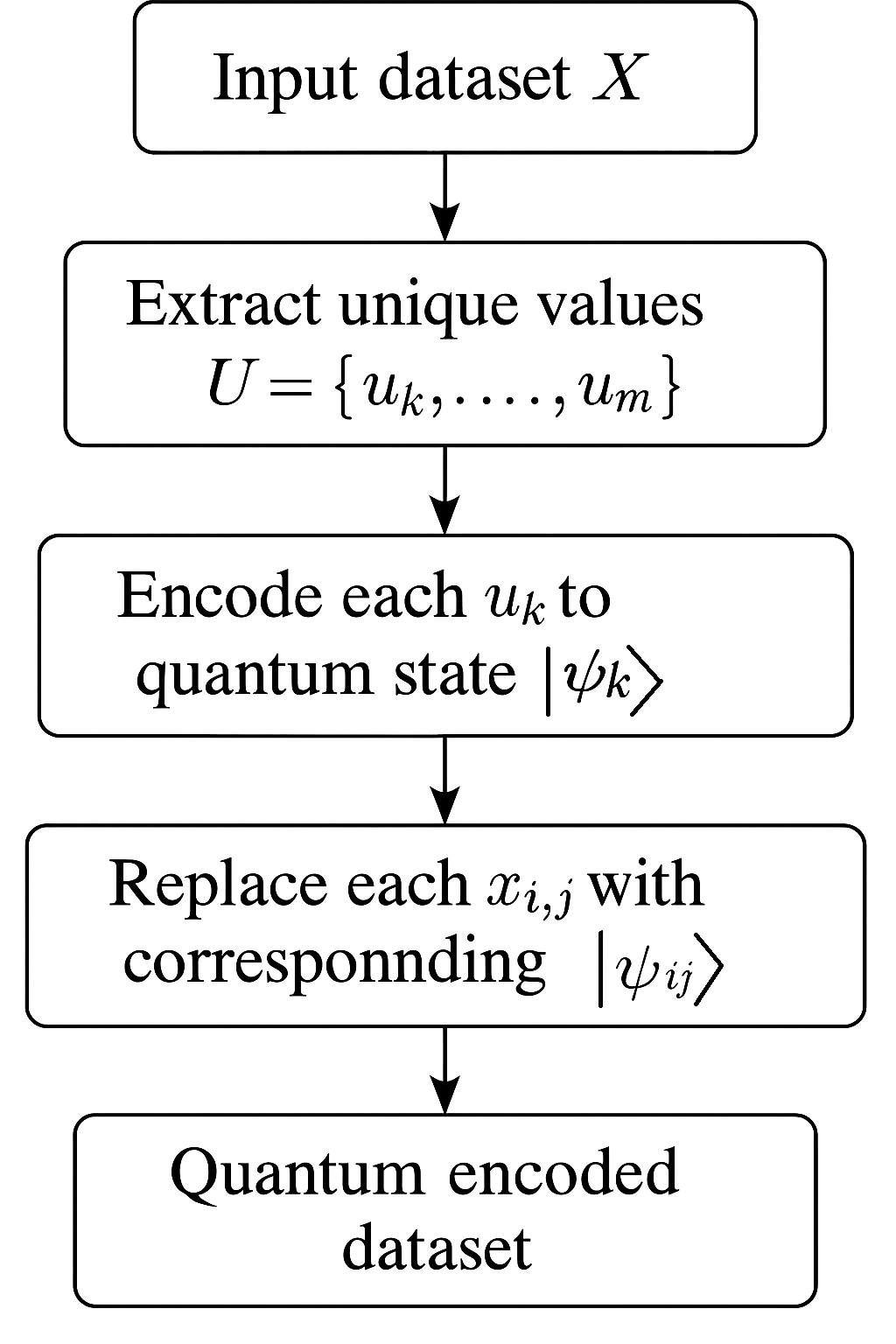}
		\caption{Flow Chart for Instance level strategy}
		\label{fig:GDSFlowChart}
	\end{center}
\end{wrapfigure}

\noindent As shown in Figure \ref{fig:GDSFlowChart} ,this transformation ensures that repeated values across different instances are encoded once, significantly reducing redundancy. When reconstructing the dataset, each classical feature value is replaced with its quantum representation. This method reduces the number of encoding operations, thereby improving computational efficiency. However, the loss of row  specific variation means that different instances may share the same quantum encoding, potentially impacting model expressiveness. The complexity is approximately $O(m)$, where $m$ is the number of unique values across all features. Since $m$ depends on the distinct values in the feature space, its magnitude relative to $n$ is not determined ($n.d.$), meaning it can be smaller, equal to, or even greater than $n$ depending on the dataset.\\
For Example: Consider the following dataset:
\[
X = 
\begin{bmatrix}
	0.2 & 0.2 & 0.9 \\
	0.4 & 0.4 & 0.1 \\
	0.6 & 0.6 & 0.3 \\
	0.2 & 0.4 & 0.3
\end{bmatrix}
\]

Step 1: Identify all unique values across the dataset:
\[
U = \{0.1,\ 0.2,\ 0.3,\ 0.4,\ 0.6,\ 0.9\}
\]

Step 2: Apply quantum encoding to each unique value:
\[
\begin{aligned}
	E(0.1) &\rightarrow |\psi_1\rangle , E(0.2) &\rightarrow |\psi_2\rangle, 
	E(0.3) &\rightarrow |\psi_3\rangle, 
	E(0.4) &\rightarrow |\psi_4\rangle, 
	E(0.6) &\rightarrow |\psi_5\rangle ,
	E(0.9) &\rightarrow |\psi_6\rangle
\end{aligned}
\]

Step 3: Reconstruct the dataset by replacing each value with its corresponding quantum state:
\[
X_{\text{quantum}} = 
\begin{bmatrix}
	|\psi_2\rangle & |\psi_2\rangle & |\psi_6\rangle \\
	|\psi_4\rangle & |\psi_4\rangle & |\psi_1\rangle \\
	|\psi_5\rangle & |\psi_5\rangle & |\psi_3\rangle \\
	|\psi_2\rangle & |\psi_4\rangle & |\psi_3\rangle
\end{bmatrix}
\]

\subsection{Class Conditional Value Strategy}

Class Conditional Value Strategy is a quantum encoding strategy that leverages class label information to guide the representation of feature values. Unlike global strategies that treat all data uniformly, this approach ensures that feature values are encoded in a manner sensitive to their associated class. This retains class-specific structural patterns in the quantum representation, thereby enhancing the model's ability to distinguish between classes.\\

This strategy can be further divided into two distinct types based on the granularity of encoding:

\begin{itemize}
	\item[] \text{Class Conditional Instance Level Strategy ( CC-ILS)}: In this approach, each instance is independently encoded within its respective class. This preserves both row specific and class specific variation, allowing fine grained representation that reflects intra class diversity.
	
	\item[] \text{Class Conditional Global Discrete Strategy ( CC-GDS)}: Here, all unique values within each class are identified and encoded once. These encodings are then reused across all instances belonging to that class. This reduces redundancy while still preserving class wise differences in feature representations.
\end{itemize}

\subsection{Class Conditional Instance Level Strategy ( CC-ILS)}
	
This strategy combines the strengths of instance level and class aware encoding. Each data instance is transformed into a quantum state, but the encoding is done with reference to its class membership. This ensures that both the instance specific variation and class specific semantics are preserved in the quantum representation. The method is particularly beneficial for classification tasks where intra class patterns are significant.

Let $X = \{x_i\}$ be the dataset and $Y = \{y_i\}$ be the corresponding class labels with $c$ unique classes. Each instance $x_i$ is encoded using an embedding function $E_y$ specific to its class $y_i$:

\begin{equation}
	E_{y_i}(x_i) \to |\psi_i^{y_i}\rangle
\end{equation}

\noindent where $|\psi_i^{y_i}\rangle$ is the quantum state of instance $x_i$ encoded using the class specific function $E_{y_i}$. This process results in $n$ unique quantum states, one for each instance, with encoding rules adapted to each class.

The complexity of this strategy is:
\[
O(n \cdot d \cdot C_{\text{embed}}^{\text{class}})
\]
where $C_{\text{embed}}^{\text{class}}$ denotes the cost of embedding a single feature value using class specific rules. This cost varies depending on both the encoding method (e.g., angle, basis, amplitude) and the class conditioned transformation logic.

\begin{wrapfigure}{r}{0.6\textwidth}
	\begin{center}
		\includegraphics[width=0.9\linewidth]{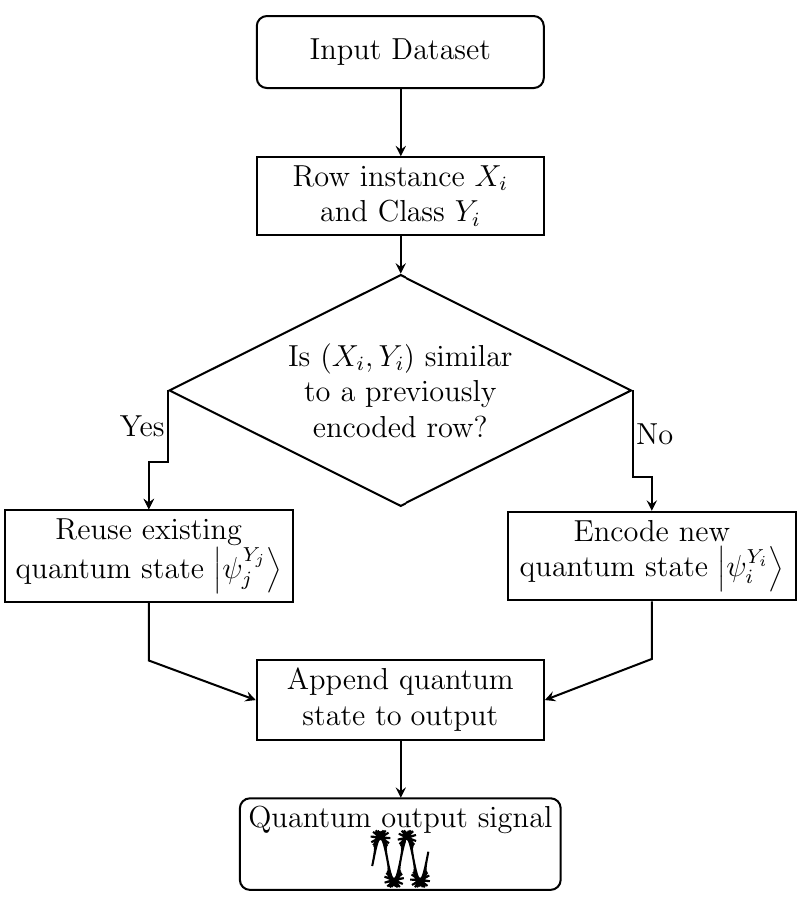}
		\caption{Flow Chart for Class Conditional Instance Level Strategy}
		\vspace*{-0.5em}
		\label{fig:CCIILSFlowChart}
	\end{center}
\end{wrapfigure}

While this method incurs a high computational cost due to the instance wise transformation, it ensures high representational fidelity within each class. Like standard ILS, this method can incorporate redundancy reduction mechanisms by reusing encodings for identical or similar instances within the same class.

\paragraph{Example:} Consider a dataset with two classes and instance level variation:

\[
X =
\begin{bmatrix}
	\text{Class} & f_1 & f_2 \\
	\hline
	A & 0.2 & 0.3 \\
	A & 0.4 & 0.3 \\
	B & 0.2 & 0.1 \\
	B & 0.2 & 0.1 \\
\end{bmatrix}
\]

Since rows 3 and 4 belong to the same class and are identical, they may share the same quantum encoding:

\[
\begin{aligned}
	E_A([0.2, 0.3]) &\to |\psi_1^A\rangle \\
	E_A([0.4, 0.3]) &\to |\psi_2^A\rangle \\
	E_B([0.2, 0.1]) &\to |\psi_1^B\rangle \\
	E_B([0.2, 0.1]) &\to |\psi_1^B\rangle
\end{aligned}
\]

\[
X_{\text{quantum}} =
\begin{bmatrix}
	A & |\psi_1^A\rangle \\
	A & |\psi_2^A\rangle \\
	B & |\psi_1^B\rangle \\
	B & |\psi_1^B\rangle
\end{bmatrix}
\]

This strategy enhances intra class expressiveness, enabling finer discrimination of class specific instance patterns while supporting redundancy aware optimization.

\subsection{ Class Conditional Global Discrete Strategy ( CC-GDS) }
\begin{wrapfigure}{r}{0.5\textwidth}
	\begin{center}
		\includegraphics[width=0.6\linewidth]{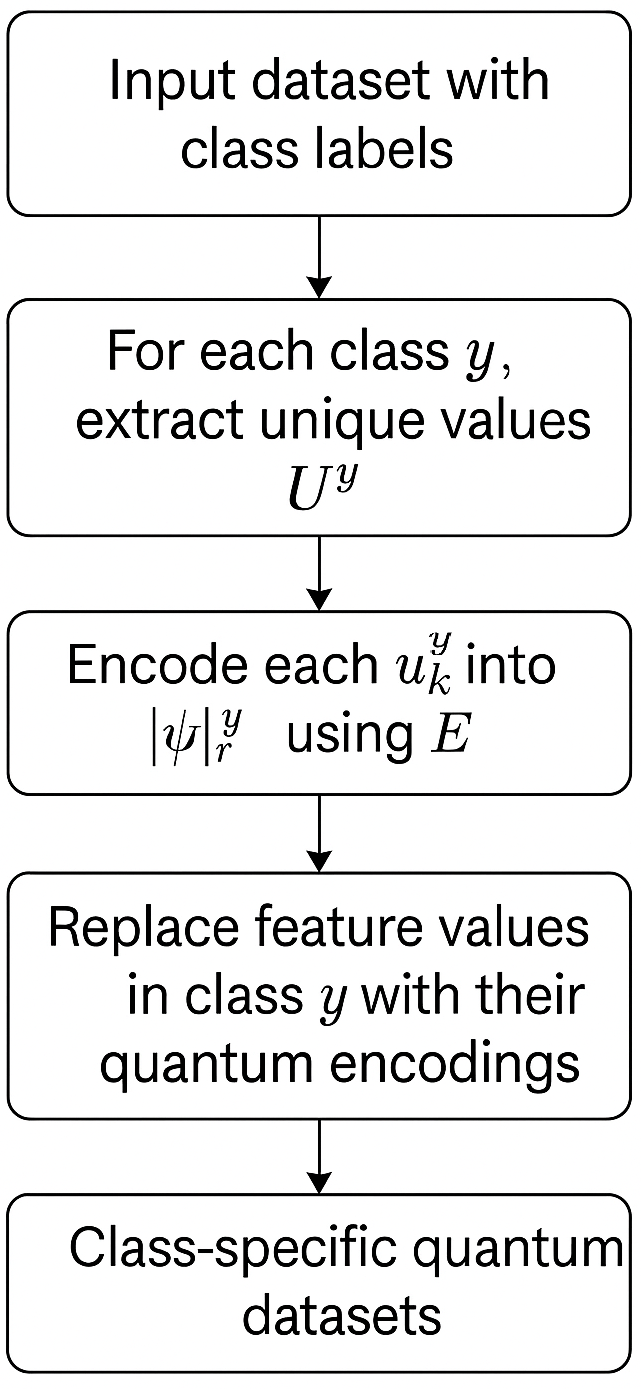}
		\caption{Flow Chart for Class Conditional Global Discrete Strategy ( CC-GDS)}
		\label{fig:CLSFlowChart}
	\end{center}
\end{wrapfigure}

Class Conditional Global Strategy ( CC-GDS) as shown in Figure \ref{fig:CLSFlowChart} extends the global approach by ensuring that unique feature values are encoded separately for each class. This preserves class  dependent variations while still benefiting from reduced redundancy. For a dataset with class labels $Y = \{y_1, y_2, \dots, y_c\}$, where $c$ is the number of unique classes, define $U_y = \{u_1^y, u_2^y, \dots, u_m^y\}$ as the set of unique values appearing within class $y$. The encoding steps are:

\begin{enumerate}
	\item Extract Unique Values Per Class: Identify the set of unique feature values appearing in each class separately.
	\item Quantum Encoding: Apply the encoding function $E$ to each unique value in the class:
	\begin{equation}
		E(u_k^y) \to |\psi_k^y\rangle, \quad \forall k,y
	\end{equation}
	\item Reconstruct Class  Specific Quantum Dataset: Each feature value in a given class is replaced with its corresponding quantum representation.
\end{enumerate}

\noindent Unlike global encoding, which applies a single mapping across the entire dataset, class  wise encoding ensures that values within different classes receive different quantum states. This helps retain class  specific data patterns and improves classification performance while still reducing redundancy. The complexity is approximately $O(c \cdot m)$, where $c$ is the number of classes and $m$ is the average number of unique values per class. While this is higher than global encoding, it provides better classification performance due to class  dependent feature retention. By balancing redundancy reduction with class level information retention, this method offers a middle ground between row  wise and global encoding strategies.\\

for Example: 

Consider a dataset with two classes and the following feature values:

\[
X = 
\begin{bmatrix}
	\text{Class} & f_1 & f_2 \\
	\hline
	A & 0.2 & 0.3 \\
	A & 0.4 & 0.3 \\
	B & 0.2 & 0.1 \\
	B & 0.6 & 0.1 \\
\end{bmatrix}
\]

Step 1: Extract unique values per class:

\[
\begin{aligned}
	U^A &= \{0.2, 0.3, 0.4\} \\
	U^B &= \{0.1, 0.2, 0.6\}
\end{aligned}
\]

Step 2: Encode values class wise:
\[
\begin{aligned}
	E^A(0.2) &\to |\psi_1^A\rangle, \quad E^A(0.3) \to |\psi_2^A\rangle, \quad E^A(0.4) \to |\psi_3^A\rangle \\
	E^B(0.1) &\to |\psi_1^B\rangle, \quad E^B(0.2) \to |\psi_2^B\rangle, \quad E^B(0.6) \to |\psi_3^B\rangle
\end{aligned}
\]

Step 3: Reconstruct class specific quantum datasets:

\[
X_{\text{quantum}} =
\begin{bmatrix}
	\text{Class} & f_1 & f_2 \\
	\hline
	A & |\psi_1^A\rangle & |\psi_2^A\rangle \\
	A & |\psi_3^A\rangle & |\psi_2^A\rangle \\
	B & |\psi_2^B\rangle & |\psi_1^B\rangle \\
	B & |\psi_3^B\rangle & |\psi_1^B\rangle \\
\end{bmatrix}
\]

\section{Implementation}

We implemented a set of quantum inspired data encoding strategies and experimented with multiple quantum data embedding techniques using the open source PennyLane library. Our study spans both Discrete Quantum Computing (DQC) and Continuous Variable Quantum Computing (CVQC) embedding paradigms, applied to a customer churn classification problem in the telecommunications domain. \\

The dataset, publicly available on Kaggle, contains 7,043 customer records with 20 features and a binary target indicating churn. Categorical features include gender, SeniorCitizen, Partner, Dependents, PhoneService, MultipleLines, InternetService, OnlineSecurity, OnlineBackup, DeviceProtection, TechSupport, StreamingTV, StreamingMovies, Contract, PaperlessBilling, PaymentMethod, and Churn. Numerical features include tenure, MonthlyCharges, and TotalCharges. Due to a strong correlation (0.83) between tenure and TotalCharges, the latter was excluded. Features like PhoneService (VIF $\approx$ 12) and MonthlyCharges (VIF $\approx$ 6) were also removed to reduce multicollinearity. The customerID feature was dropped as it held no predictive value.

\begin{wrapfigure}{r}{0.7\textwidth}
	\begin{center}
		\includegraphics[width=0.6\linewidth]{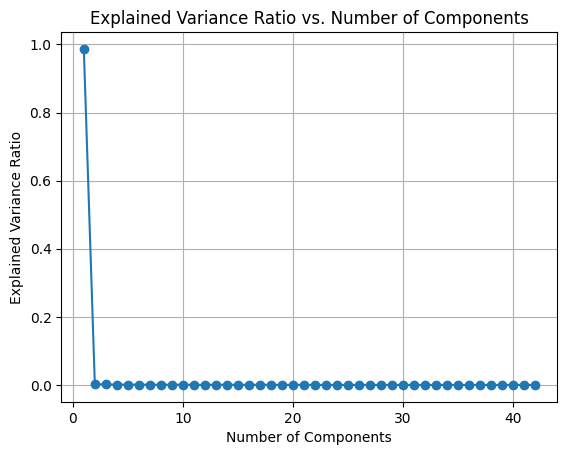}
		\caption{Explained Variance Ratio with Elbow Point}
		\label{fig:ExplainedVarianceElbow}
	\end{center}
	\begin{quote}
		\centering
		Elbow Point Index: 23\
		Explained Variance Ratio at Elbow Point: 8.126234391114803e  33 \
		Cumulative Explained Variance at Elbow Point: 1.0000000000000002 \
	\end{quote}
\end{wrapfigure}

After preprocessing and one-hot encoding of categorical variables, the dataset expanded to 42 features. Due to severe class imbalance (1,869 positive cases vs. 5,174 negative cases), we retained all minority class instances and randomly undersampled the majority class. This resulted in a balanced dataset of 3,738 samples.

To determine the optimal number of principal components, we performed PCA and plotted the explained variance ratio. As shown in Figure~\ref{fig:ExplainedVarianceElbow}, the elbow point was identified at 23 components, which we retained for downstream quantum encoding.

We explored six types of quantum embeddings. These belong to two categories. The first category is direct encoding methods that use discrete variable qubit based representations such as Basis, Angle, Instantaneous Quantum Polynomial-time (IQP), and Quantum Approximate Optimization Algorithm (QAOA). The QAOA is a variational Quantum Approximate Optimization Algorithm implemented on qubit systems \cite{farhi2014quantumapproximateoptimizationalgorithm} \cite{lloyd2018quantumapproximateoptimizationcomputationally}. The second category is continuous variable encoding methods represented by Displacement and Squeezing embeddings.  Displacement Embedding  
encodes data by shifting the quantum state in phase space using the displacement operator: $  D(\alpha) = \exp\left(\alpha \hat{a}^\dagger - \alpha^* \hat{a}\right) $.  This typically maps classical data to the mean of a Gaussian wavefunction, producing a coherent state \( \ket{\alpha} = D(\alpha)\ket{0} \) \cite{killoran2019continuous}. Squeezing Embedding encodes data by altering the uncertainty (variance) of a quantum state using the squeezing operator: $ S(r) = \exp\left[\frac{1}{2} r \left( \hat{a}^2 - \hat{a}^{\dagger 2} \right) \right] $.  This adjusts the shape (spread) of the wavefunction in phase space \cite{lloyd1999quantum} \cite{weedbrook2012gaussian}.

Each embedding type was implemented using four encoding strategies.

Direct Encoding (DE): A baseline with no specific strategy applied.
Instance Level Strategy (ILS): Encodes each row individually based on its feature values.
Global Discrete Strategy (GDS): Uses the globally unique set of values across the dataset.
Class Conditional Instance Level Strategy (CC-ILS): Encodes rows separately for each class label (churned vs. not churned) using ILS.

Table~\ref{tab:EncodingTime} summarizes the encoding time for each strategy, highlighting the computational cost associated with each strategy and technique.

\begin{table}[htbp]
	\centering
	\caption{Encoding Time (in Seconds) for Different Quantum-Inspired Encoding Strategies}		\label{tab:EncodingTime} 
	\begin{tabular}{@{}lll@{}}
		\toprule
		\textbf{Encoding Type} & \textbf{Encoding Strategy} & \textbf{Encoding Time (s)} \\
		\midrule
		\multirow{4}{*}{Basis} 
		& DE (3738 Rows)                     & 8189.8900 \\
		&  ILS (2951 Rows)                     & 4862.9500 \\
		&  GDS (2 values)                     & 0.0000 \\
		& CC-ILS (0- 1804 Rows, 1- 1690 Rows)            & (3123.9154 + 2912.9060) = 6036.8296 \\
		\midrule
		\multirow{4}{*}{Angle} 
		& DE                                 & 7877.3700 \\
		&  ILS                                 & 6086.5100 \\
		&  GDS (76194 Unique Values)          & 27.1041 \\
		& CC-ILS (0-1804 Rows, 1-1690 Rows)     & (3889.0186 + 3514.7746) = 7403.7962 \\
		\midrule
		\multirow{4}{*}{IQP} 
		& DE                                 & 33676.4006 \\
		&  ILS (3455 Rows)                     & 28288.3900 \\
		&  GDS (76947 Values)                 & 37.2156 \\
		& CC-ILS (0-1806 Rows, 1-1690 Rows)     & (15991.9950 + 15269.9551) = 31261.9501 \\
		\midrule
		\multirow{4}{*}{Displacement} 
		& DE                                 & 41.4430 \\
		&  ILS (3432 Rows)                     & 20.8100 \\
		&  GDS (72662 Values)                 & 63.3668 \\
		& CC-ILS (0-1811 Rows, 1-1690 Rows)     & 20.7454 \\
		\midrule
		\multirow{4}{*}{Squeezing} 
		& DE                                 & 52.3153 \\
		&  ILS (3437 Rows)                     & 15.7600 \\
		&  GDS (74215 Values)                 & 66.9308 \\
		& CC-ILS (0-1791 Rows, 1-1690 Rows)     & 9.9500 \\
		\midrule
		\multirow{4}{*}{QAOA} 
		& DE                                 & 63505.5914 \\
		&  ILS (3448 Rows)                     & 67171.5900 \\
		&  GDS (74489 Values)                 & 166.7059 \\
		& CC-ILS (0-1810 Rows, 1-1690 Rows)     & (32810.1705 + 30026.8884) = 62836.0589 \\
		\bottomrule
	\end{tabular}
\end{table}

The results in table \ref{tab:EncodingTime}highlight clear trends regarding computational efficiency and scalability of each encoding approach.\\\\ Among the encoding types, QAOA and IQP exhibit substantially higher encoding times compared to the other methods, with DE times of approximately 63,506 seconds and 33,676 seconds respectively. In contrast, Displacement and Squeezing encodings demonstrate markedly lower computational overhead, requiring less than 60 seconds for DE. The Basis and Angle encodings fall between these extremes, with DE times around 8,000 seconds.\\
Regarding encoding strategies, the GDS consistently yields the lowest encoding times across all encoding types, ranging from near zero for Basis to under 170 seconds for QAOA. This efficiency is attributable to the strategy’s reliance on encoding unique values rather than the entire dataset, resulting in significant computational savings. ILS also generally reduces encoding time relative to DE by focusing on a subset of the data, though the QAOA encoding shows an exception where ILS incurs slightly higher time, potentially due to implementation overhead or complexity.\\
The CC ILS strategy, which performs encoding separately on class specific subsets, generally incurs a combined encoding time comparable to or slightly less than the sum of the individual DE encoding times for each class. This approach offers marginal computational benefits, particularly for Displacement and Squeezing encodings, where CC-ILS encoding times fall below 21 seconds.\\
Overall, these results suggest that encoding type is the dominant factor influencing encoding time, with QAOA and IQP requiring substantially more computational resources. Strategies that leverage data reduction, such as GDS and ILS, can provide meaningful time savings and improve scalability. The choice of encoding strategy should thus consider the trade off between computational efficiency and potential impacts on model performance.\\
For modeling, we employed various classical and quantum machine learning approaches. Classical models included Logistic Regression, K  Nearest Neighbors (KNN), Support Vector Machines (SVM), and ensemble methods such as Random Forest, LightGBM, AdaBoost, and CatBoost. Logistic Regression was used as a baseline model due to its interpretability and effectiveness in binary classification. KNN was included for its instance  based learning approach, while SVM was selected for its ability to handle complex decision boundaries. The ensemble models were incorporated to leverage multiple weak learners and enhance predictive performance.\\
Logistic Regression is a linear model widely used for binary classification tasks due to its simplicity, interpretability, and probabilistic output. It estimates the probability that a given input belongs to a particular class using the logistic sigmoid function\cite{hosmer2013applied}. \\
K-Nearest Neighbors (KNN) is a non-parametric, instance-based learning algorithm that classifies new samples based on the majority class among the k nearest data points in the feature space. It is known for its simplicity and effectiveness in low-dimensional problems \cite{cover1967nearest}.\\
Support Vector Machines (SVM) are supervised learning models that find the optimal hyperplane to separate data points of different classes by maximizing the margin between them. Kernel methods extend SVM to non-linear decision boundaries, making it suitable for high-dimensional and complex datasets \cite{cortes1995support}.\\
Random Forest is an ensemble learning technique that builds multiple decision trees using bootstrapped datasets and random feature selection. It aggregates their predictions through majority voting, improving robustness and reducing overfitting \cite{breiman2001random}.\\
LightGBM is a gradient boosting framework that uses histogram-based algorithms and leaf-wise tree growth, which significantly enhances training speed and model accuracy on large datasets \cite{ke2017lightgbm}.\\
AdaBoost (Adaptive Boosting) sequentially trains weak classifiers, typically decision stumps, focusing on misclassified examples in subsequent iterations to reduce bias and variance \cite{freund1997decision}.\\
CatBoost is a gradient boosting algorithm designed to handle categorical features efficiently without the need for extensive preprocessing. It employs ordered boosting and symmetric tree structures to reduce overfitting and training time \cite{prokhorenkova2018catboost}.

\begin{landscape}
	\begin{longtable}{|p{1.5cm}|p{1.9cm}|p{1.6cm}|p{1.9cm}|p{1.5cm}|p{1.5cm}|p{1.0cm}|p{1.0cm}|p{1.0cm}|p{1.3cm}|p{1.5cm}|}
		\caption{Classical data, Discrete and Continuous Variable Quantum Data Embedding Performance} 
		\label{tab:Results} \\
		\hline
		\textbf{Classifier}  & \textbf{Encoding Type} & \textbf{Encoding Strategy} & \textbf{Encoding Time (In Seconds)
		}&\textbf{Accuracy} & \textbf{Precision}  & \textbf{Recall} & \textbf{F1 Score} & \textbf{ROC AUC} & \textbf{Cohen's Kappa} & \textbf{Running Time} \\
		\hline
		\endfirsthead
		\multicolumn{11}{c}%
		{{\tablename\ \thetable{} -- continued from previous page}} \\
		\hline
	\textbf{Classifier}  & \textbf{Encoding Type} & \textbf{Encoding Strategy} & \textbf{Encoding Time (In Seconds)
	}&\textbf{Accuracy} & \textbf{Precision}  & \textbf{Recall} & \textbf{F1 Score} & \textbf{ROC AUC} & \textbf{Cohen's Kappa} & \textbf{Running Time} \\
		\hline
		\endhead
		\hline \multicolumn{11}{|r|}{{Continued on next page}} \\ \hline
		\endfoot
		
		\hline
		\endlastfoot
		
\multirow{24}{*}{\makecell[{{p{1.5cm}}}]{Logistic \\ Regression}} 
& \multirow{4}{*}{\makecell[{{p{1.0cm}}}]{ Basis}}
& DE& 8189.8900 & 0.6684 & 0.6512 & 0.7440 & 0.6945 & 0.6674 & 0.3354 & 0.0061 \\
&&  ILS & 4862.9500 & 0.7433 & 0.7431 & 0.7433 & 0.7432 & 0.8042 & 0.4864 & 0.0077 \\
&&  GDS & 0.0000 & 0.7401 & 0.7399 & 0.7396 & 0.7397 & 0.8030 & 0.4795 & 0.0076 \\
&& CC-ILS & 6036.8296 & 0.7251 & 0.7250 & 0.7252 & 0.7250 & 0.7891 & 0.4501 & 0.0112 \\
\cline{2-11}
&\multirow{4}{*}{\makecell[{{p{1.0cm}}}]{ Angle}}
& DE& 7877.3700 & 0.6684 & 0.6512 & 0.7440 & 0.6945 & 0.6674 & 0.3354 & 0.0029 \\
&&  ILS & 6086.5100 & 0.7251 & 0.7250 & 0.7252 & 0.7250 & 0.7904 & 0.4501 & 0.0250 \\
&&  GDS & 27.1041 & 0.7487 & 0.7491 & 0.7493 & 0.7487 & 0.7988 & 0.4977 & 0.0000 \\
&& CC-ILS & 7403.7962 & 0.7401 & 0.7404 & 0.7406 & 0.7401 & 0.7995 & 0.4804 & 0.0120 \\
\cline{2-11}
&\multirow{4}{*}{\makecell[{{p{1.0cm}}}]{IQP}}
& DE& 33676.40063 & 49.3316 & 0.0000 & 0.0000 & 0.0000 & 0.5000 & 0.0000 & 0.0340 \\
&&  ILS & 28288.3900 & 0.4845 & 0.2422 & 0.5000 & 0.3264 & 0.5000 & 0.0000 & 0.0036 \\
&&  GDS & 37.2156 & 0.4845 & 0.2422 & 0.5000 & 0.3264 & 0.5000 & 0.0000 & 0.0020 \\
&& CC-ILS & 31261.9501 & 0.4845 & 0.2422 & 0.5000 & 0.3264 & 0.5000 & 0.0000 & 0.0073 \\
\cline{2-11}
&\multirow{4}{*}{\makecell[{{p{1.0cm}}}]{Displacement}}
& DE& 41.4430 & 0.7259 & 0.7254 & 0.7388 & 0.7320 & 0.7258 & 0.4516 & 0.0676 \\
&&  ILS & 20.8100 & 0.6332 & 0.6327 & 0.6323 & 0.6323 & 0.6979 & 0.2649 & 0.0233 \\
&&  GDS & 63.3668 & 0.7091 & 0.7131 & 0.7109 & 0.7087 & 0.7807 & 0.4201 & 0.0607 \\
&& CC-ILS & 20.7454 & 0.6513 & 0.6510 & 0.6509 & 0.6510 & 0.6930 & 0.3019 & 0.0364 \\
\cline{2-11}
&\multirow{4}{*}{\makecell[{{p{1.0cm}}}]{Sqeezing}}
& DE& 52.3153 & 0.7326 & 0.7361 & 0.7361 & 0.7361 & 0.7326 & 0.4651 & 0.0129 \\
&&  ILS & 15.7600 & 0.7551 & 0.7557 & 0.7539 & 0.7542 & 0.8285 & 0.5088 & 0.0767 \\
&&  GDS & 66.9308 & 0.7230 & 0.7227 & 0.7227 & 0.7227 & 0.7909 & 0.4454 & 0.0799 \\
&& CC-ILS & 9.9500 & 0.7615 & 0.7613 & 0.7611 & 0.7612 & 0.8288 & 0.5224 & 0.0567 \\
\cline{2-11}&\multirow{4}{*}{\makecell[{{p{1.0cm}}}]{QAOA}}
& DE& 63505.5914 & 0.5120 & 0.5207 & 0.4644 & 0.4909 & 0.5127 & 0.0253 & 0.0050 \\
&&  ILS & 67171.5900 & 0.4909 & 0.4918 & 0.4918 & 0.4908 & 0.4991 & 0.0163 & 0.0050 \\
&&  GDS & 166.7059 & 0.5326 & 0.5351 & 0.5346 & 0.5316 & 0.5461 & 0.0689 & 0.0216 \\
&& CC-ILS & 62836.0589 & 0.5059 & 0.5080 & 0.5079 & 0.5047 & 0.5186 & 0.0157 & 0.0051 \\
\cline{2-11}
\hline
\multirow{4}{*}{\makecell[{{p{1.5cm}}}]{KNN}}
& \multirow{4}{*}{\makecell[{{p{1.0cm}}}]{ Basis}}
& DE& 8189.8900 & 0.5067 & 0.5067 & 1.0000 & 0.6726 & 0.5000 & 0.0000 & 0.0884 \\
&&  ILS & 4862.9500 & 0.6941 & 0.6941 & 0.6931 & 0.6932 & 0.7591 & 0.3868 & 0.0091 \\
&&  GDS & 0.0000 & 0.7102 & 0.7104 & 0.7091 & 0.7092 & 0.7609 & 0.4188 & 0.0083 \\
&& CC-ILS & 6036.8296 & 0.6684 & 0.6689 & 0.6668 & 0.6666 & 0.7278 & 0.3345 & 0.2254 \\
\cline{2-11}
\pagebreak
\multirow{20}{*}{\makecell[{{p{1.5cm}}}]{KNN continued..}}
&\multirow{4}{*}{\makecell[{{p{1.0cm}}}]{ Angle}}
& DE& 7877.3700 & 0.5067 & 0.5067 & 1.0000 & 0.6726 & 0.5000 & 0.0000 & 0.0448 \\
&&  ILS & 6086.5100 & 0.7070 & 0.7110 & 0.7043 & 0.7036 & 0.7536 & 0.4106 & 0.0140 \\
&&  GDS & 27.1041 & 0.7091 & 0.7100 & 0.7075 & 0.7076 & 0.7759 & 0.4161 & 0.0090 \\
&& CC-ILS & 7403.7962 & 0.6930 & 0.6954 & 0.6908 & 0.6903 & 0.7631 & 0.3831 & 0.1779 \\
\cline{2-11}
&\multirow{4}{*}{\makecell[{{p{1.0cm}}}]{IQP}}
& DE& 33676.40063 & 62.0321 & 0.6323 & 0.5989 & 0.6152 & 0.6206 & 0.2410 & 0.2960 \\
&&  ILS & 28288.3900 & 0.5230 & 0.5233 & 0.5233 & 0.5230 & 0.5353 & 0.0466 & 0.1478 \\
&&  GDS & 37.2156 & 0.5155 & 0.2578 & 0.5000 & 0.3402 & 0.5000 & 0.0000 & 0.2062 \\
&& CC-ILS & 31261.9501 & 0.4866 & 0.4873 & 0.4873 & 0.4866 & 0.4988 & -0.0254 & 0.0114 \\
\cline{2-11}
&\multirow{4}{*}{\makecell[{{p{1.0cm}}}]{Displacement}}
& DE& 41.4430 & 0.7112 & 0.7139 & 0.7177 & 0.7158 & 0.7111 & 0.4223 & 0.0995 \\
&&  ILS & 20.8100 & 0.6888 & 0.6885 & 0.6886 & 0.6886 & 0.7580 & 0.3771 & 0.0128 \\
&&  GDS & 63.3668 & 0.7337 & 0.7335 & 0.7332 & 0.7333 & 0.7847 & 0.4666 & 0.2135 \\
&& CC-ILS & 20.7454 & 0.6684 & 0.6681 & 0.6678 & 0.6679 & 0.7249 & 0.3358 & 0.2952 \\
\cline{2-11}
&\multirow{4}{*}{\makecell[{{p{1.0cm}}}]{Squeezing}}
& DE& 52.3153 & 0.7072 & 0.7073 & 0.7203 & 0.7137 & 0.7070 & 0.4142 & 0.0154 \\
&&  ILS & 15.7600 & 0.7037 & 0.7066 & 0.7014 & 0.7010 & 0.7553 & 0.4045 & 0.1569 \\
&&  GDS & 66.9308 & 0.7048 & 0.7059 & 0.7031 & 0.7031 & 0.7603 & 0.4074 & 0.2120 \\
&& CC-ILS & 9.9500 & 0.7316 & 0.7349 & 0.7293 & 0.7292 & 0.7753 & 0.4605 & 0.1813 \\
\cline{2-11}&\multirow{4}{*}{\makecell[{{p{1.0cm}}}]{QAOA}}
& DE& 63505.5914 & 0.4813 & 0.4858 & 0.4063 & 0.4425 & 0.4823 & -0.0353 & 0.0080 \\
&&  ILS & 67171.5900 & 0.5316 & 0.5327 & 0.5326 & 0.5314 & 0.5355 & 0.0649 & 0.0092 \\
&&  GDS & 166.7059 & 0.5433 & 0.5438 & 0.5438 & 0.5433 & 0.5480 & 0.0874 & 0.2134 \\
&& CC-ILS & 62836.0589 & 0.5626 & 0.5625 & 0.5626 & 0.5624 & 0.5646 & 0.1250 & 0.0085 \\
\cline{2-11}
\hline
\multirow{8}{*}{\makecell[{{p{1.5cm}}}]{SVM \\ Linear}}
& \multirow{4}{*}{\makecell[{{p{1.0cm}}}]{ Basis}}
& DE& 8189.8900 & 0.6684 & 0.6513 & 0.7441 & 0.6946 & 0.6674 & 0.3355 & 0.5428 \\
&&  ILS & 4862.9500 & 0.7209 & 0.7273 & 0.7232 & 0.7201 & 0.7932 & 0.4441 & 0.9130 \\
&&  GDS & 0.0000 & 0.7198 & 0.7248 & 0.7218 & 0.7192 & 0.7958 & 0.4416 & 0.9059 \\
&& CC-ILS & 6036.8296 & 0.7155 & 0.7220 & 0.7179 & 0.7147 & 0.7857 & 0.4335 & 2.0414 \\
\cline{2-11}
&\multirow{4}{*}{\makecell[{{p{1.0cm}}}]{ Angle}}
& DE& 7877.3700 & 66.8449 & 0.6513 & 0.7441 & 0.6946 & 0.6674 & 0.3355 & 0.1568 \\
&&  ILS & 6086.5100 & 0.7358 & 0.7357 & 0.7359 & 0.7357 & 0.7885 & 0.4715 & 1.2550 \\
&&  GDS & 27.1041 & 0.7497 & 0.7499 & 0.7501 & 0.7497 & 0.7978 & 0.4996 & 0.7593 \\
&& CC-ILS & 7403.7962 & 0.7316 & 0.7316 & 0.7319 & 0.7315 & 0.7950 & 0.4632 & 1.2759 \\
\cline{2-11}
\pagebreak
\multirow{16}{*}{\makecell[{{p{1.5cm}}}]{SVM \\ Linear continued..}}
&\multirow{4}{*}{\makecell[{{p{1.0cm}}}]{IQP}}
& DE& 33676.40063 & 49.3316 & 0.0000 & 0.0000 & 0.0000 & 0.5000 & 0.0000 & 0.3130 \\
&&  ILS & 28288.3900 & 0.4845 & 0.2422 & 0.5000 & 0.3264 & 0.5000 & 0.0000 & 0.8867 \\
&&  GDS & 37.2156 & 0.4845 & 0.2422 & 0.5000 & 0.3264 & 0.5000 & 0.0000 & 1.0962 \\
&& CC-ILS & 31261.9501 & 0.4845 & 0.2422 & 0.5000 & 0.3264 & 0.5000 & 0.0000 & 1.0252 \\
\cline{2-11}
&\multirow{4}{*}{\makecell[{{p{1.0cm}}}]{Displacement}}
& DE& 41.4430 & 0.7139 & 0.7538 & 0.6464 & 0.6960 & 0.7148 & 0.4288 & 0.8031 \\
&&  ILS & 20.8100 & 0.6267 & 0.6265 & 0.6253 & 0.6251 & 0.6981 & 0.2511 & 1.2578 \\
&&  GDS & 63.3668 & 0.7144 & 0.7246 & 0.7175 & 0.7128 & 0.7812 & 0.4321 & 1.1026 \\
&& CC-ILS & 20.7454 & 0.6503 & 0.6499 & 0.6494 & 0.6494 & 0.6974 & 0.2991 & 1.3567 \\
\cline{2-11}
&\multirow{4}{*}{\makecell[{{p{1.0cm}}}]{Sqeezing}}
& DE& 52.3153 & 0.7139 & 0.7538 & 0.6464 & 0.6960 & 0.7148 & 0.4288 & 0.3902 \\
&&  ILS & 15.7600 & 0.7326 & 0.7379 & 0.7300 & 0.7295 & 0.8236 & 0.4622 & 1.0530 \\
&&  GDS & 66.9308 & 0.7369 & 0.7389 & 0.7351 & 0.7352 & 0.7942 & 0.4717 & 2.8696 \\
&& CC-ILS & 9.9500 & 0.7658 & 0.7670 & 0.7645 & 0.7647 & 0.8349 & 0.5301 & 1.2910 \\
\cline{2-11}
&\multirow{4}{*}{\makecell[{{p{1.0cm}}}]{QAOA}}
& DE& 63505.5914 & 0.5254 & 0.5357 & 0.4749 & 0.5035 & 0.5261 & 0.0521 & 0.2361 \\
&&  ILS & 67171.5900 & 0.4963 & 0.4973 & 0.4973 & 0.4961 & 0.5042 & -0.0053 & 0.8363 \\
&&  GDS & 166.7059 & 0.5326 & 0.5351 & 0.5346 & 0.5316 & 0.5482 & 0.0689 & 1.4317 \\
&& CC-ILS & 62836.0589 & 0.5123 & 0.5133 & 0.5133 & 0.5122 & 0.4811 & 0.0265 & 0.8629 \\
\cline{2-11}
\hline
\multirow{12}{*}{\makecell[{{p{1.5cm}}}]{SVM \\ Poly}} 
& \multirow{4}{*}{\makecell[{{p{1.0cm}}}]{ Basis}}
& DE& 8189.8900 & 0.6684 & 0.6513 & 0.7441 & 0.6946 & 0.6674 & 0.3355 & 0.7717 \\
&&  ILS & 4862.9500 & 0.7144 & 0.7142 & 0.7141 & 0.7141 & 0.7852 & 0.4283 & 0.7470 \\
&&  GDS & 0.0000 & 0.7144 & 0.7142 & 0.7143 & 0.7142 & 0.7825 & 0.4284 & 0.7433 \\
&& CC-ILS & 6036.8296 & 0.7144 & 0.7146 & 0.7134 & 0.7136 & 0.7804 & 0.4275 & 1.5852 \\
\cline{2-11}
&\multirow{4}{*}{\makecell[{{p{1.0cm}}}]{ Angle}}
& DE& 7877.3700 & 0.6684 & 0.6513 & 0.7441 & 0.6946 & 0.6674 & 0.3355 & 0.1801 \\
&&  ILS & 6086.5100 & 0.7144 & 0.7163 & 0.7125 & 0.7124 & 0.7704 & 0.4265 & 1.7001 \\
&&  GDS & 27.1041 & 0.7390 & 0.7397 & 0.7378 & 0.7380 & 0.7930 & 0.4765 & 1.1971 \\
&& CC-ILS & 7403.7962 & 0.7176 & 0.7178 & 0.7166 & 0.7167 & 0.7808 & 0.4338 & 2.0159 \\
\cline{2-11}
&\multirow{4}{*}{\makecell[{{p{1.0cm}}}]{IQP}}
& DE& 33676.40063 & 0.6043 & 0.5719 & 0.8707 & 0.6904 & 0.6007 & 0.2028 & 0.2929 \\
&&  ILS & 28288.3900 & 0.4930 & 0.5009 & 0.5007 & 0.4647 & 0.5190 & 0.0013 & 1.0042 \\
&&  GDS & 37.2156 & 0.4845 & 0.2422 & 0.5000 & 0.3264 & 0.5000 & 0.0000 & 1.1728 \\
&& CC-ILS & 31261.9501 & 0.4963 & 0.4999 & 0.4999 & 0.4909 & 0.4874 & -0.0002 & 1.1936 \\
\cline{2-11}
\pagebreak
\multirow{12}{*}{\makecell[{{p{1.5cm}}}]{SVM \\ Poly Continued ..}} 
&\multirow{4}{*}{\makecell[{{p{1.0cm}}}]{Displacement}}
& DE& 41.4430 & 0.7099 & 0.6875 & 0.7836 & 0.7324 & 0.7089 & 0.4186 & 42.9444 \\
&&  ILS & 20.8100 & 0.6898 & 0.6910 & 0.6907 & 0.6898 & 0.7407 & 0.3806 & 10073.0766 \\
&&  GDS & 63.3668 & 0.6984 & 0.6981 & 0.6978 & 0.6979 & 0.7589 & 0.3958 & 10910.5044 \\
&& CC-ILS & 20.7454 & 0.6738 & 0.6735 & 0.6733 & 0.6733 & 0.7234 & 0.3467 & 9776.6768 \\
\cline{2-11}
&\multirow{4}{*}{\makecell[{{p{1.0cm}}}]{Sqeezing}}
& DE& 52.3153 & 0.7380 & 0.7328 & 0.7599 & 0.7461 & 0.7377 & 0.4756 & 0.5365 \\
&&  ILS & 15.7600 & 0.7176 & 0.7215 & 0.7152 & 0.7147 & 0.7790 & 0.4323 & 6452.0129 \\
&&  GDS & 66.9308 & 0.7091 & 0.7094 & 0.7079 & 0.7080 & 0.7704 & 0.4165 & 3703.7335 \\
&& CC-ILS & 9.9500 & 0.7262 & 0.7265 & 0.7251 & 0.7253 & 0.7882 & 0.4510 & 7491.4499 \\
\cline{2-11}
&\multirow{4}{*}{\makecell[{{p{1.0cm}}}]{QAOA}}
& DE& 63505.5914 & 0.5080 & 0.5169 & 0.4433 & 0.4773 & 0.5089 & 0.0178 & 0.2596 \\
&&  ILS & 67171.5900 & 0.5198 & 0.5206 & 0.5206 & 0.5198 & 0.5388 & 0.0410 & 0.9860 \\
&&  GDS & 166.7059 & 0.5187 & 0.5193 & 0.5193 & 0.5187 & 0.5138 & 0.0385 & 1.2623 \\
&& CC-ILS & 62836.0589 & 0.5380 & 0.5404 & 0.5398 & 0.5371 & 0.5594 & 0.0793 & 1.0093 \\
\cline{2-11}
\hline
\multirow{16}{*}{\makecell[{{p{1.5cm}}}]{SVM \\ RBF}} 
& \multirow{4}{*}{\makecell[{{p{1.0cm}}}]{ Basis}}
& DE& 8189.8900 & 0.6684 & 0.6513 & 0.7441 & 0.6946 & 0.6674 & 0.3355 & 0.9918 \\
&&  ILS & 4862.9500 & 0.7305 & 0.7304 & 0.7298 & 0.7299 & 0.8038 & 0.4599 & 1.1163 \\
&&  GDS & 0.0000 & 0.7369 & 0.7372 & 0.7358 & 0.7361 & 0.8078 & 0.4724 & 1.1111 \\
&& CC-ILS & 6036.8296 & 0.7358 & 0.7362 & 0.7347 & 0.7349 & 0.7972 & 0.4703 & 2.4398 \\
\cline{2-11}
&\multirow{4}{*}{\makecell[{{p{1.0cm}}}]{ Angle}}
& DE& 7877.3700 & 0.6684 & 0.6513 & 0.7441 & 0.6946 & 0.6674 & 0.3355 & 0.2802 \\
&&  ILS & 6086.5100 & 0.7390 & 0.7394 & 0.7380 & 0.7382 & 0.7970 & 0.4767 & 1.4245 \\
&&  GDS & 27.1041 & 0.7572 & 0.7571 & 0.7574 & 0.7571 & 0.8140 & 0.5143 & 1.1099 \\
&& CC-ILS & 7403.7962 & 0.7487 & 0.7485 & 0.7487 & 0.7485 & 0.8016 & 0.4971 & 1.9364 \\
\cline{2-11}
&\multirow{4}{*}{\makecell[{{p{1.0cm}}}]{IQP}}
& DE& 33676.40063 & 0.6364 & 0.6126 & 0.7678 & 0.6815 & 0.6346 & 0.2701 & 0.4064 \\
&&  ILS & 28288.3900 & 0.4877 & 0.4903 & 0.4907 & 0.4843 & 0.5278 & -0.0185 & 1.6125 \\
&&  GDS & 37.2156 & 0.4845 & 0.2422 & 0.5000 & 0.3264 & 0.5000 & 0.0000 & 1.6422 \\
&& CC-ILS & 31261.9501 & 0.4834 & 0.4882 & 0.4911 & 0.4543 & 0.5046 & -0.0176 & 1.9518 \\
\cline{2-11}
&\multirow{4}{*}{\makecell[{{p{1.0cm}}}]{Displacement}}
& DE& 41.4430 & 0.7460 & 0.7455 & 0.7573 & 0.7513 & 0.7458 & 0.4918 & 0.4938 \\
&&  ILS & 20.8100 & 0.7337 & 0.7342 & 0.7325 & 0.7327 & 0.8053 & 0.4659 & 1.1391 \\
&&  GDS & 63.3668 & 0.7294 & 0.7293 & 0.7287 & 0.7289 & 0.7968 & 0.4578 & 1.9497 \\
&& CC-ILS & 20.7454 & 0.7615 & 0.7614 & 0.7610 & 0.7611 & 0.8206 & 0.5222 & 1.3585 \\
\cline{2-11}
\pagebreak
\multirow{4}{*}{\makecell[{{p{1.5cm}}}]{SVM \\ RBF Continued..}} 
&\multirow{4}{*}{\makecell[{{p{1.0cm}}}]{QAOA}}
& DE& 63505.5914 & 0.5147 & 0.5263 & 0.4222 & 0.4685 & 0.5160 & 0.0318 & 0.3820 \\
&&  ILS & 67171.5900 & 0.5519 & 0.5533 & 0.5531 & 0.5517 & 0.5679 & 0.1058 & 1.4834 \\
&&  GDS & 166.7059 & 0.5722 & 0.5738 & 0.5734 & 0.5720 & 0.6032 & 0.1464 & 2.1994 \\
&& CC-ILS & 62836.0589 & 0.5369 & 0.5373 & 0.5373 & 0.5369 & 0.5463 & 0.0746 & 1.4696 \\
\cline{2-11}
\hline
\multirow{24}{*}{\makecell[{{p{1.5cm}}}]{SVM \\ Sigmoid}} 
& \multirow{4}{*}{\makecell[{{p{1.0cm}}}]{ Basis}}
& DE& 8189.8900 & 0.6684 & 0.6513 & 0.7441 & 0.6946 & 0.6674 & 0.3355 & 0.0017 \\
&&  ILS & 4862.9500 & 0.6471 & 0.6486 & 0.6482 & 0.6470 & 0.7096 & 0.2955 & 0.9096 \\
&&  GDS & 0.0000 & 0.6770 & 0.6771 & 0.6773 & 0.6770 & 0.7336 & 0.3542 & 0.9109 \\
&& CC-ILS & 6036.8296 & 0.6588 & 0.6607 & 0.6601 & 0.6587 & 0.7070 & 0.3191 & 1.6452 \\
\cline{2-11}
&\multirow{4}{*}{\makecell[{{p{1.0cm}}}]{ Angle}}
& DE& 7877.3700 & 0.6684 & 0.6513 & 0.7441 & 0.6946 & 0.6674 & 0.3355 & 0.0012 \\
&&  ILS & 6086.5100 & 0.3540 & 0.3542 & 0.3541 & 0.3540 & 0.3208 & -0.2913 & 1.1252 \\
&&  GDS & 27.1041 & 0.3818 & 0.3823 & 0.3829 & 0.3814 & 0.3367 & -0.2333 & 0.7614 \\
&& CC-ILS & 7403.7962 & 0.3594 & 0.3598 & 0.3602 & 0.3592 & 0.3094 & -0.2787 & 1.0961 \\
\cline{2-11}
&\multirow{4}{*}{\makecell[{{p{1.0cm}}}]{IQP}}
& DE& 33676.40063 & 0.5214 & 0.5279 & 0.5251 & 0.5265 & 0.5213 & 0.0427 & 0.2880 \\
&&  ILS & 28288.3900 & 0.5166 & 0.5163 & 0.5163 & 0.5163 & 0.5231 & 0.0326 & 1.0111 \\
&&  GDS & 37.2156 & 0.4845 & 0.2422 & 0.5000 & 0.3264 & 0.5000 & 0.0000 & 1.1798 \\
&& CC-ILS & 31261.9501 & 0.5037 & 0.5039 & 0.5039 & 0.5037 & 0.5082 & 0.0077 & 1.2031 \\
\cline{2-11}
&\multirow{4}{*}{\makecell[{{p{1.0cm}}}]{Displacement}}
& DE& 41.4430 & 0.4933 & 0.0000 & 0.0000 & 0.0000 & 0.5000 & 0.0000 & 0.4181 \\
&&  ILS & 20.8100 & 0.4845 & 0.2422 & 0.5000 & 0.3264 & 0.5000 & 0.0000 & 1.7806 \\
&&  GDS & 63.3668 & 0.4845 & 0.2422 & 0.5000 & 0.3264 & 0.5000 & 0.0000 & 1.8489 \\
&& CC-ILS & 20.7454 & 0.4845 & 0.2422 & 0.5000 & 0.3264 & 0.5000 & 0.0000 & 0.9429 \\
\cline{2-11}
&\multirow{4}{*}{\makecell[{{p{1.0cm}}}]{Sqeezing}}
& DE& 52.3153 & 0.5548 & 0.5599 & 0.5673 & 0.5636 & 0.5546 & 0.1093 & 0.5428 \\
&&  ILS & 15.7600 & 0.4845 & 0.2422 & 0.5000 & 0.3264 & 0.5000 & 0.0000 & 1.0383 \\
&&  GDS & 66.9308 & 0.4845 & 0.2422 & 0.5000 & 0.3264 & 0.5000 & 0.0000 & 2.0252 \\
&& CC-ILS & 9.9500 & 0.4845 & 0.2422 & 0.5000 & 0.3264 & 0.5000 & 0.0000 & 1.1757 \\
\cline{2-11}&\multirow{4}{*}{\makecell[{{p{1.0cm}}}]{QAOA}}
& DE& 63505.5914 & 0.5040 & 0.5118 & 0.4565 & 0.4826 & 0.5047 & 0.0093 & 0.3687 \\
&&  ILS & 67171.5900 & 0.4909 & 0.4915 & 0.4915 & 0.4909 & 0.4993 & -0.0170 & 1.3401 \\
&&  GDS & 166.7059 & 0.5219 & 0.5251 & 0.5245 & 0.5198 & 0.4804 & 0.0487 & 1.7730 \\
&& CC-ILS & 62836.0589 & 0.4684 & 0.4692 & 0.4693 & 0.4684 & 0.5243 & -0.0612 & 1.3663 \\
\cline{2-11}
\hline
\pagebreak
\multirow{24}{*}{\makecell[{{p{1.5cm}}}]{Decision Tree}} 
 & \multirow{4}{*}{\makecell[{{p{1.0cm}}}]{ Basis}}
& DE& 8189.89 & 0.6684 & 0.6513 & 0.7441 & 0.6946 & 0.6674 & 0.3355 & 0.0017 \\
&&  ILS & 4862.95 & 0.6385 & 0.6399 & 0.6395 & 0.6384 & 0.6475 & 0.2783 & 0.006 \\
&&  GDS & 0.0 & 0.6717 & 0.6725 & 0.6724 & 0.6717 & 0.6797 & 0.3441 & 0.0069 \\
&& CC-ILS & 6036.8296 & 0.6342 & 0.6339 & 0.6339 & 0.6339 & 0.6415 & 0.2678 & 0.0301 \\
\cline{2-11}
&\multirow{4}{*}{\makecell[{{p{1.0cm}}}]{ Angle}}
& DE& 7877.37 & 0.6684 & 0.6513 & 0.7441 & 0.6946 & 0.6674 & 0.3355 & 0.0012 \\
&&  ILS & 6086.51 & 0.6663 & 0.6662 & 0.6663 & 0.6662 & 0.6693 & 0.3324 & 0.1065 \\
&&  GDS & 27.1041 & 0.6941 & 0.6948 & 0.6948 & 0.6941 & 0.6986 & 0.3889 & 0.0652 \\
&& CC-ILS & 7403.7962 & 0.6791 & 0.6788 & 0.6787 & 0.6787 & 0.682 & 0.3575 & 0.1147 \\
\cline{2-11}
&\multirow{4}{*}{\makecell[{{p{1.0cm}}}]{IQP}}
& DE& 33676.40063 & 0.6203 & 0.6294 & 0.6095 & 0.6193 & 0.6205 & 0.2408 & 0.1180 \\
&&  ILS & 28288.3900 & 0.5561 & 0.5567 & 0.5567 & 0.5561 & 0.5540 & 0.1132 & 0.1186 \\
&&  GDS & 37.2156 & 0.4845 & 0.2422 & 0.5000 & 0.3264 & 0.5000 & 0.0000 & 0.0020 \\
&& CC-ILS & 31261.9501 & 0.5273 & 0.5266 & 0.5266 & 0.5266 & 0.5185 & 0.0533 & 0.0971 \\
\cline{2-11}
&\multirow{4}{*}{\makecell[{{p{1.0cm}}}]{Displacement}}
& DE& 41.4430 & 0.6858 & 0.7057 & 0.6517 & 0.6776 & 0.6863 & 0.3722 & 0.0181 \\
&&  ILS & 20.8100 & 0.6364 & 0.6360 & 0.6360 & 0.6360 & 0.6377 & 0.2720 & 0.0920 \\
&&  GDS & 63.3668 & 0.6481 & 0.6504 & 0.6495 & 0.6479 & 0.6498 & 0.2980 & 0.0965 \\
&& CC-ILS & 20.7454 & 0.6599 & 0.6599 & 0.6601 & 0.6598 & 0.6642 & 0.3199 & 0.0580 \\
\cline{2-11}
&\multirow{4}{*}{\makecell[{{p{1.0cm}}}]{Sqeezing}}
& DE& 52.3153 & 0.6832 & 0.6994 & 0.6570 & 0.6776 & 0.6835 & 0.3667 & 0.0110 \\
&&  ILS & 15.7600 & 0.6663 & 0.6660 & 0.6656 & 0.6657 & 0.6759 & 0.3314 & 0.0846 \\
&&  GDS & 66.9308 & 0.6802 & 0.6800 & 0.6801 & 0.6800 & 0.6888 & 0.3601 & 0.1094 \\
&& CC-ILS & 9.9500 & 0.6684 & 0.6681 & 0.6680 & 0.6680 & 0.6651 & 0.3361 & 0.0797 \\
\cline{2-11}&\multirow{4}{*}{\makecell[{{p{1.0cm}}}]{QAOA}}
& DE& 63505.5914 & 0.5267 & 0.5369 & 0.4802 & 0.5070 & 0.5274 & 0.0547 & 0.0932 \\
&&  ILS & 67171.5900 & 0.5508 & 0.5501 & 0.5500 & 0.5499 & 0.5400 & 0.1000 & 0.0943 \\
&&  GDS & 166.7059 & 0.5316 & 0.5318 & 0.5318 & 0.5315 & 0.5298 & 0.0636 & 0.1123 \\
&& CC-ILS & 62836.0589 & 0.5594 & 0.5600 & 0.5600 & 0.5594 & 0.5600 & 0.1197 & 0.0600 \\
\cline{2-11}
\hline
\multirow{6}{*}{\makecell[{{p{1.5cm}}}]{Random \\ Forest}}
& \multirow{4}{*}{\makecell[{{p{1.0cm}}}]{ Basis}}
& DE& 8189.8900 & 0.6684 & 0.6513 & 0.7441 & 0.6946 & 0.6674 & 0.3355 & 1.1630 \\
&&  ILS & 4862.9500 & 0.7102 & 0.7102 & 0.7104 & 0.7101 & 0.7995 & 0.4204 & 0.2208 \\
&&  GDS & 0.0000 & 0.7155 & 0.7154 & 0.7156 & 0.7154 & 0.7964 & 0.4309 & 0.2208 \\
&& CC-ILS & 6036.8296 & 0.6930 & 0.6927 & 0.6927 & 0.6927 & 0.7670 & 0.3855 & 0.4948 \\
\cline{2-11}
&\multirow{2}{*}{\makecell[{{p{1.0cm}}}]{ Angle}}
& DE& 7877.3700 & 0.6684 & 0.6513 & 0.7441 & 0.6946 & 0.6674 & 0.3355 & 0.1917 \\
&&  ILS & 6086.5100 & 0.7422 & 0.7422 & 0.7415 & 0.7417 & 0.8051 & 0.4835 & 1.0745 \\
\pagebreak
\multirow{18}{*}{\makecell[{{p{1.5cm}}}]{Random \\ Forest Continued..}}
&\multirow{2}{*}{\makecell[{{p{1.0cm}}}]{ Angle}}
&  GDS & 27.1041 & 0.7369 & 0.7369 & 0.7371 & 0.7368 & 0.8155 & 0.4738 & 0.7881 \\
&& CC-ILS & 7403.7962 & 0.7412 & 0.7412 & 0.7414 & 0.7411 & 0.8066 & 0.4823 & 1.2243 \\
\cline{2-11}
&\multirow{4}{*}{\makecell[{{p{1.0cm}}}]{IQP}}
& DE& 33676.40063 & 0.6524 & 0.6639 & 0.6359 & 0.6496 & 0.6526 & 0.3051 & 0.1141 \\
&&  ILS & 28288.3900 & 0.5765 & 0.5785 & 0.5779 & 0.5761 & 0.5993 & 0.1553 & 1.0800 \\
&&  GDS & 37.2156 & 0.4845 & 0.2422 & 0.5000 & 0.3264 & 0.5000 & 0.0000 & 0.1297 \\
&& CC-ILS & 31261.9501 & 0.5123 & 0.5127 & 0.5127 & 0.5123 & 0.5248 & 0.0253 & 1.2302 \\
\cline{2-11}
&\multirow{4}{*}{\makecell[{{p{1.0cm}}}]{Displacement}}
& DE& 41.4430 & 0.7193 & 0.7183 & 0.7335 & 0.7258 & 0.7191 & 0.4382 & 0.0956 \\
&&  ILS & 20.8100 & 0.7412 & 0.7438 & 0.7426 & 0.7410 & 0.8094 & 0.4836 & 1.1386 \\
&&  GDS & 63.3668 & 0.7380 & 0.7387 & 0.7387 & 0.7380 & 0.8194 & 0.4764 & 1.1680 \\
&& CC-ILS & 20.7454 & 0.7241 & 0.7256 & 0.7251 & 0.7240 & 0.7897 & 0.4491 & 0.7608 \\
\cline{2-11}
&\multirow{4}{*}{\makecell[{{p{1.0cm}}}]{Sqeezing}}
& DE& 52.3153 & 0.7193 & 0.7183 & 0.7335 & 0.7258 & 0.7191 & 0.4382 & 0.1211 \\
&&  ILS & 15.7600 & 0.7294 & 0.7292 & 0.7289 & 0.7290 & 0.7998 & 0.4581 & 0.8318 \\
&&  GDS & 66.9308 & 0.7262 & 0.7262 & 0.7264 & 0.7261 & 0.8012 & 0.4524 & 1.2545 \\
&& CC-ILS & 9.9500 & 0.7326 & 0.7324 & 0.7326 & 0.7325 & 0.8068 & 0.4649 & 0.9863 \\
\cline{2-11}&\multirow{4}{*}{\makecell[{{p{1.0cm}}}]{QAOA}}
& DE& 63505.5914 & 0.5334 & 0.5403 & 0.5303 & 0.5353 & 0.5335 & 0.0669 & 0.0608 \\
&&  ILS & 67171.5900 & 0.5283 & 0.5292 & 0.5291 & 0.5283 & 0.5706 & 0.0581 & 0.9662 \\
&&  GDS & 166.7059 & 0.5626 & 0.5660 & 0.5648 & 0.5613 & 0.6089 & 0.1289 & 1.3178 \\
&& CC-ILS & 62836.0589 & 0.5561 & 0.5590 & 0.5581 & 0.5552 & 0.6090 & 0.1157 & 0.9140 \\
\cline{2-11}
\hline
\multirow{12}{*}{\makecell[{{p{1.5cm}}}]{AdaBoost}} 
& \multirow{4}{*}{\makecell[{{p{1.0cm}}}]{ Basis}}
&DE&8189.8900 &0.6684 &0.6513 &0.7441 &0.6946 &0.6674 &0.3355 &0.5223\\
&& ILS &4862.9500 &0.7390 &0.7390 &0.7392 &0.7390 &0.8031 &0.4780 &0.0974\\
&& GDS &0.0000 &0.7358 &0.7356 &0.7355 &0.7355 &0.8006 &0.4710 &0.0955\\
&&CC-ILS &6036.8296 &0.7198 &0.7198 &0.7200 &0.7197 &0.7843 &0.4396 &0.1910\\
\cline{2-11}
&\multirow{4}{*}{\makecell[{{p{1.0cm}}}]{ Angle}}
&DE&7877.3700 &0.6684 &0.6513 &0.7441 &0.6946 &0.6674 &0.3355 &0.1311\\
&& ILS &6086.5100 &0.7390 &0.7392 &0.7381 &0.7383 &0.8056 &0.4769 &0.4750\\
&& GDS &27.1041 &0.7636 &0.7646 &0.7624 &0.7627 &0.8317 &0.5258 &0.3716\\
&&CC-ILS &7403.7962 &0.7412 &0.7409 &0.7409 &0.7409 &0.8199 &0.4819 &0.5555\\
\cline{2-11}
&\multirow{4}{*}{\makecell[{{p{1.0cm}}}]{IQP}}
&DE&33676.40063 &67.2460 &0.6516 &0.7599 &0.7016 &0.6713 &0.3433 &0.4047\\
&& ILS &28288.3900 &0.4995 &0.5016 &0.5016 &0.4981 &0.5001 &0.0031 &0.4095\\
&& GDS &37.2156 &0.4845 &0.2422 &0.5000 &0.3264 &0.5000 &0.0000 &0.0060\\
&&CC-ILS &31261.9501 &0.4834 &0.4888 &0.4942 &0.4187 &0.5048 &-0.0112 &0.4753\\
\cline{2-11}
\pagebreak
\multirow{12}{*}{\makecell[{{p{1.5cm}}}]{AdaBoost Continued..}} 
&\multirow{4}{*}{\makecell[{{p{1.0cm}}}]{Displacement}}
&DE&41.4430 &0.7286 &0.7316 &0.7335 &0.7325 &0.7285 &0.4571 &0.1799\\
&& ILS &20.8100 &0.6952 &0.6969 &0.6963 &0.6951 &0.7769 &0.3915 &0.5517\\
&& GDS &63.3668 &0.7251 &0.7287 &0.7268 &0.7249 &0.8181 &0.4519 &0.5194\\
&&CC-ILS &20.7454 &0.7326 &0.7334 &0.7334 &0.7326 &0.7916 &0.4658 &0.3782\\
\cline{2-11}
&\multirow{4}{*}{\makecell[{{p{1.0cm}}}]{Sqeezing}}
&DE&52.3153 &0.7286 &0.7316 &0.7335 &0.7325 &0.7285 &0.4571 &0.1830\\
&& ILS &15.7600 &0.7358 &0.7368 &0.7344 &0.7346 &0.8095 &0.4699 &0.4140\\
&& GDS &66.9308 &0.7348 &0.7347 &0.7340 &0.7342 &0.8134 &0.4685 &0.6381\\
&&CC-ILS &9.9500 &0.7572 &0.7580 &0.7560 &0.7563 &0.8240 &0.5130 &0.5207\\
\cline{2-11}&\multirow{4}{*}{\makecell[{{p{1.0cm}}}]{QAOA}}
&DE&63505.5914 &0.5334 &0.5560 &0.3931 &0.4606 &0.5353 &0.0704 &0.4172\\
&& ILS &67171.5900 &0.5262 &0.5303 &0.5292 &0.5231 &0.5581 &0.0581 &0.3835\\
&& GDS &166.7059 &0.5668 &0.5693 &0.5686 &0.5662 &0.6068 &0.1366 &0.6001\\
&&CC-ILS &62836.0589 &0.5358 &0.5346 &0.5343 &0.5338 &0.5435 &0.0688 &0.3796\\
\cline{2-11}
\hline
\multirow{16}{*}{\makecell[{{p{1.5cm}}}]{Extra \\ Trees}}
& \multirow{4}{*}{\makecell[{{p{1.0cm}}}]{ Basis}}
&DE&8189.8900 &0.6684 &0.6513 &0.7441 &0.6946 &0.6674 &0.3355 &0.5852\\
&& ILS &4862.9500 &0.7123 &0.7127 &0.7128 &0.7123 &0.7776 &0.4249 &0.2492\\
&& GDS &0.0000 &0.7070 &0.7073 &0.7074 &0.7069 &0.7766 &0.4142 &0.2469\\
&&CC-ILS &6036.8296 &0.6824 &0.6821 &0.6821 &0.6821 &0.7503 &0.3641 &0.4454\\
\cline{2-11}
&\multirow{4}{*}{\makecell[{{p{1.0cm}}}]{ Angle}}
&DE&7877.3700 &0.6684 &0.6513 &0.7441 &0.6946 &0.6674 &0.3355 &0.1351\\
&& ILS &6086.5100 &0.7112 &0.7110 &0.7111 &0.7110 &0.7804 &0.4221 &0.3618\\
&& GDS &27.1041 &0.7283 &0.7286 &0.7287 &0.7283 &0.8010 &0.4569 &0.2839\\
&&CC-ILS &7403.7962 &0.7348 &0.7347 &0.7349 &0.7347 &0.7856 &0.4694 &0.4253\\
\cline{2-11}
&\multirow{4}{*}{\makecell[{{p{1.0cm}}}]{IQP}}
&DE&33676.40063 &0.6497 &0.6552 &0.6517 &0.6534 &0.6497 &0.2994 &0.3730\\
&& ILS &28288.3900 &0.5551 &0.5587 &0.5574 &0.5535 &0.5850 &0.1142 &0.3530\\
&& GDS &37.2156 &0.4845 &0.2422 &0.5000 &0.3264 &0.5000 &0.0000 &0.1018\\
&&CC-ILS &31261.9501 &0.5380 &0.5405 &0.5399 &0.5370 &0.5392 &0.0794 &0.4163\\
\cline{2-11}
&\multirow{4}{*}{\makecell[{{p{1.0cm}}}]{Displacement}}
&DE&41.4430 &0.7219 &0.7355 &0.7045 &0.7197 &0.7222 &0.4441 &0.4927\\
&& ILS &20.8100 &0.7102 &0.7152 &0.7122 &0.7096 &0.7805 &0.4225 &0.3979\\
&& GDS &63.3668 &0.7209 &0.7234 &0.7223 &0.7207 &0.7973 &0.4431 &0.4648\\
&&CC-ILS &20.7454 &0.7112 &0.7138 &0.7127 &0.7111 &0.7716 &0.4239 &0.2850\\
\cline{2-11}
\pagebreak
\multirow{8}{*}{\makecell[{{p{1.5cm}}}]{Extra \\ Trees Continued..}}
&\multirow{4}{*}{\makecell[{{p{1.0cm}}}]{Sqeezing}}
&DE&52.3153 &0.7273 &0.7384 &0.7150 &0.7265 &0.7274 &0.4547 &0.4869\\
&& ILS &15.7600 &0.7027 &0.7025 &0.7027 &0.7026 &0.7845 &0.4052 &0.3097\\
&& GDS &66.9308 &0.7070 &0.7070 &0.7072 &0.7069 &0.7778 &0.4140 &0.5107\\
&&CC-ILS &9.9500 &0.7326 &0.7324 &0.7324 &0.7324 &0.7888 &0.4647 &0.3673\\
\cline{2-11}&\multirow{4}{*}{\makecell[{{p{1.0cm}}}]{QAOA}}
&DE&63505.5914 &0.5174 &0.5262 &0.4776 &0.5007 &0.5179 &0.0358 &0.3390\\
&& ILS &67171.5900 &0.5316 &0.5334 &0.5331 &0.5310 &0.5678 &0.0659 &0.3186\\
&& GDS &166.7059 &0.5679 &0.5732 &0.5708 &0.5654 &0.6137 &0.1407 &0.3751\\
&&CC-ILS &62836.0589 &0.5668 &0.5720 &0.5697 &0.5644 &0.6169 &0.1385 &0.3146\\
\cline{2-11}
\hline
\multirow{20}{*}{\makecell[{{p{1.5cm}}}]{Gradient \\ Boosting}}
& \multirow{4}{*}{\makecell[{{p{1.0cm}}}]{ Basis}}
&DE&8189.8900 &0.6684 &0.6513 &0.7441 &0.6946 &0.6674 &0.3355 &0.1531\\
&& ILS &4862.9500 &0.7316 &0.7317 &0.7306 &0.7308 &0.8155 &0.4618 &0.2280\\
&& GDS &0.0000 &0.7358 &0.7360 &0.7349 &0.7351 &0.8246 &0.4704 &0.2216\\
&&CC-ILS &6036.8296 &0.7412 &0.7415 &0.7401 &0.7403 &0.8110 &0.4810 &0.3998\\
\cline{2-11}
&\multirow{4}{*}{\makecell[{{p{1.0cm}}}]{ Angle}}
&DE&7877.3700 &0.6684 &0.6513 &0.7441 &0.6946 &0.6674 &0.3355 &0.1027\\
&& ILS &6086.5100 &0.7476 &0.7494 &0.7460 &0.7461 &0.8182 &0.4933 &2.3009\\
&& GDS &27.1041 &0.7604 &0.7608 &0.7595 &0.7597 &0.8390 &0.5197 &1.6919\\
&&CC-ILS &7403.7962 &0.7615 &0.7617 &0.7606 &0.7609 &0.8289 &0.5219 &2.2226\\
\cline{2-11}
&\multirow{4}{*}{\makecell[{{p{1.0cm}}}]{IQP}}
&DE&33676.40063 &0.6618 &0.6591 &0.6887 &0.6735 &0.6614 &0.3230 &2.0365\\
&& ILS &28288.3900 &0.5241 &0.5265 &0.5261 &0.5229 &0.5332 &0.0519 &1.7943\\
&& GDS &37.2156 &0.4845 &0.2422 &0.5000 &0.3264 &0.5000 &0.0000 &0.0830\\
&&CC-ILS &31261.9501 &0.5102 &0.5105 &0.5105 &0.5101 &0.5223 &0.0209 &2.0910\\
\cline{2-11}
&\multirow{4}{*}{\makecell[{{p{1.0cm}}}]{Displacement}}
&DE&41.4430 &0.7487 &0.7468 &0.7625 &0.7546 &0.7485 &0.4971 &0.4710\\
&& ILS &20.8100 &0.7262 &0.7268 &0.7269 &0.7262 &0.8064 &0.4529 &2.4164\\
&& GDS &63.3668 &0.7529 &0.7528 &0.7525 &0.7526 &0.8371 &0.5052 &2.1227\\
&&CC-ILS &20.7454 &0.7455 &0.7452 &0.7453 &0.7452 &0.8043 &0.4905 &1.6740\\
\cline{2-11}
&\multirow{4}{*}{\makecell[{{p{1.0cm}}}]{Sqeezing}}
&DE&52.3153 &0.7487 &0.7468 &0.7625 &0.7546 &0.7485 &0.4971 &0.4179\\
&& ILS &15.7600 &0.7401 &0.7412 &0.7387 &0.7389 &0.8196 &0.4785 &1.7979\\
&& GDS &66.9308 &0.7444 &0.7442 &0.7438 &0.7440 &0.8158 &0.4880 &2.6276\\
&&CC-ILS &9.9500 &0.7551 &0.7548 &0.7548 &0.7548 &0.8294 &0.5097 &2.1493\\
\cline{2-11}
\pagebreak
\multirow{4}{*}{\makecell[{{p{1.5cm}}}]{Gradient \\ Boosting Continued.. }}
&\multirow{4}{*}{\makecell[{{p{1.0cm}}}]{QAOA}}
&DE&63505.5914 &0.4906 &0.4971 &0.4485 &0.4716 &0.4912 &-0.0176 &1.8667\\
&& ILS &67171.5900 &0.5198 &0.5208 &0.5208 &0.5197 &0.5462 &0.0414 &1.6821\\
&& GDS &166.7059 &0.6011 &0.6016 &0.6016 &0.6011 &0.6380 &0.2029 &2.0882\\
&&CC-ILS &62836.0589 &0.5604 &0.5606 &0.5606 &0.5604 &0.5848 &0.1211 &1.6796\\
\cline{2-11}
\hline
\multirow{24}{*}{\makecell[{{p{1.5cm}}}]{XGBoost}}  & \multirow{4}{*}{\makecell[{{p{1.0cm}}}]{ Basis}}
&DE&8189.8900 &0.6684 &0.6513 &0.7441 &0.6946 &0.6674 &0.3355 &0.1155\\
&& ILS &4862.9500 &0.6909 &0.6910 &0.6912 &0.6909 &0.7791 &0.3819 &0.0615\\
&& GDS &0.0000 &0.6930 &0.6929 &0.6931 &0.6929 &0.7768 &0.3859 &0.0589\\
&&CC-ILS &6036.8296 &0.6920 &0.6918 &0.6911 &0.6912 &0.7601 &0.3826 &0.1628\\
\cline{2-11}
&\multirow{4}{*}{\makecell[{{p{1.0cm}}}]{ Angle}}
&DE&7877.3700 &0.6684 &0.6513 &0.7441 &0.6946 &0.6674 &0.3355 &0.0253\\
&& ILS &6086.5100 &0.7134 &0.7138 &0.7121 &0.7122 &0.7935 &0.4250 &0.1650\\
&& GDS &27.1041 &0.7358 &0.7356 &0.7355 &0.7355 &0.8137 &0.4710 &0.1082\\
&&CC-ILS &7403.7962 &0.7176 &0.7175 &0.7170 &0.7171 &0.7900 &0.4343 &0.2328\\
\cline{2-11}
&\multirow{4}{*}{\makecell[{{p{1.0cm}}}]{IQP}}
&DE&33676.40063 &0.6350 &0.6417 &0.6332 &0.6375 &0.6351 &0.2701 &0.2480\\
&& ILS &28288.3900 &0.5369 &0.5374 &0.5374 &0.5369 &0.5704 &0.0747 &0.1499\\
&& GDS &37.2156 &0.4845 &0.2422 &0.5000 &0.3264 &0.5000 &0.0000 &0.0282\\
&&CC-ILS &31261.9501 &0.5037 &0.5031 &0.5031 &0.5031 &0.5199 &0.0063 &0.1753\\
\cline{2-11}
&\multirow{4}{*}{\makecell[{{p{1.0cm}}}]{Displacement}}
&DE&41.4430 &0.7032 &0.7199 &0.6781 &0.6984 &0.7035 &0.4068 &0.0953\\
&& ILS &20.8100 &0.6995 &0.6998 &0.6999 &0.6995 &0.7883 &0.3993 &0.1833\\
&& GDS &63.3668 &0.7433 &0.7431 &0.7432 &0.7431 &0.8145 &0.4863 &0.1705\\
&&CC-ILS &20.7454 &0.7305 &0.7304 &0.7306 &0.7304 &0.7963 &0.4608 &0.1092\\
\cline{2-11}
&\multirow{4}{*}{\makecell[{{p{1.0cm}}}]{Sqeezing}}
&DE&52.3153 &0.7032 &0.7199 &0.6781 &0.6984 &0.7035 &0.4068 &0.1266\\
&& ILS &15.7600 &0.7326 &0.7330 &0.7315 &0.7317 &0.7940 &0.4638 &0.1764\\
&& GDS &66.9308 &0.7155 &0.7152 &0.7152 &0.7152 &0.7933 &0.4305 &0.2066\\
&&CC-ILS &9.9500 &0.7348 &0.7345 &0.7346 &0.7345 &0.8053 &0.4691 &0.1775\\
\cline{2-11}&\multirow{4}{*}{\makecell[{{p{1.0cm}}}]{QAOA}}
&DE&63505.5914 &0.5040 &0.5116 &0.4644 &0.4869 &0.5045 &0.0091 &0.1224\\
&& ILS &67171.5900 &0.5572 &0.5568 &0.5569 &0.5568 &0.5784 &0.1137 &0.1171\\
&& GDS &166.7059 &0.5658 &0.5677 &0.5672 &0.5654 &0.5950 &0.1339 &0.2208\\
&&CC-ILS &62836.0589 &0.5797 &0.5795 &0.5795 &0.5794 &0.6347 &0.1589 &0.1184\\
\end{longtable}
\end{landscape} 

\section*{Results and Discussion}

Our experimental results show that the encoding strategies ILS, GDS, and CCILS consistently reduce the encoding time by approximately 40 to 60\% compared to Direct Encoding (DE) across all six quantum inspired embedding methods. Despite this substantial time saving, the accuracy of classical machine learning models including SVM with linear, polynomial, and radial basis function kernels, Logistic Regression, K Nearest Neighbors, and Decision Trees trained on data encoded with these strategies remains within a small margin of variation, typically $\pm$1 to 2\%, from the accuracy achieved with DE. This demonstrates that ILS, GDS, and CCILS provide efficient alternatives for embedding classical data into quantum inspired representations, preserving model performance while greatly improving computational efficiency. These results support the use of these strategies as practical embedding methods in quantum inspired machine learning pipelines where encoding time is critical.

\section{Conclusion}
In this study, we systematically compared three quantum-inspired encoding strategies: Instance Level Strategy (ILS), Global Discrete Strategy (GDS), and Class Conditional Value Strategy (CCVS). These strategies were evaluated using six types of quantum embeddings and multiple classical classifiers. Although GDS achieved the lowest encoding time due to its value sharing mechanism, it may lead to a loss of model fidelity, particularly in large or complex datasets. On the other hand, class aware strategies such as CC-ILS, especially when combined with Squeezing encoding, delivered the highest classification accuracy by preserving critical class specific patterns. These results emphasize the importance of selecting encoding strategies based on the specific demands of the task. Encoding approaches that maintain semantic and class-level distinctions are more effective in supporting reliable and generalizable models. Future work should focus on developing hybrid or adaptive encoding strategies that balance efficiency with representational accuracy. Further exploration of variational encodings and integration with real quantum hardware is also recommended for practical scalability.

\section{Conflict of Interest Statement}
The authors declare that there is no conflict of interest regarding the publication of this manuscript.

\section*{Declarations}

\subsection*{Ethical Approval and Consent to Participate}
Not applicable. This study did not involve any human or animal subjects requiring ethical approval.

\subsection*{Consent for Publication}
Not applicable. This manuscript does not contain any individual person’s data in any form.

\subsection*{Availability of Supporting Data}
The data that support the findings of this study are available from the corresponding author upon reasonable request.

\subsection*{Competing Interests/Authors' Contributions}
The authors declare that they have no competing interests.

\textbf{Authors' Contributions:}
All authors contributed equally.

\subsection*{Funding}
This research received no specific grant from any funding agency in the public, commercial, or not  for  profit sectors.

\bibliographystyle{unsrt}
\bibliography{Row_Vs_Unique_value_Quantum_Data_Encoding}
	
\section*{Data Availability Statement}
The datasets analysed during the current study are available in the keggal repository, \\ https://www.kaggle.com/datasets/blastchar/telco-customer-churn

\section*{Author Information}

Authors and Affiliations:\\
\textbf{PhD of Analytics and Decision Science}, IIM Mumbai, India.\\
\href{mailto:minati.rath.2019@iimmumbai.co.in}{Minati Rath}.\\
\textbf{Professor of Analytics and Decision Science}, IIM Mumbai, India.\\
\href{mailto:hemadate@iimmumbai.ac.in}{Hema Date}.

\section*{Corresponding Authors}
Correspondence to \href{mailto:minati.rath.2019@iimmumbai.ac.in}{Minati Rath} or \href{mailto:hemadate@iimmumbai.ac.in}{Hema Date}.

\end{document}